\documentclass[10pt,twocolumn, hyphens]{article}
\usepackage[utf8]{inputenc}
\usepackage[T1]{fontenc}
\usepackage{amsmath,amssymb}
\usepackage{graphicx}
\usepackage{caption}
\usepackage{hyperref}
\usepackage[table]{xcolor}
\usepackage[final]{cvpr}

\definecolor{myblue}{rgb}{0.21,0.49,0.74}
\hypersetup{
    colorlinks=true,
    linkcolor=myblue,
    citecolor=myblue,
    urlcolor=myblue
}

\title{PoseAnything: Universal Pose-guided Video Generation with \\ Part-aware Temporal Coherence}

\author{
Ruiyan Wang\footnotemark[1]\thanks{Equal Contribution}
\quad Teng Hu\footnotemark[1]
\quad Kaihui Huang
\quad Zihan Su \\
\quad Ran Yi\thanks{Corresponding author.}
\quad Lizhuang Ma\\
\normalsize Shanghai Jiao Tong University\\
{\tt\small Project page: \href{https://ryan-w2024.github.io/project/PoseAnything/}{\textcolor{magenta}{https://ryan-w2024.github.io/project/PoseAnything/}}}
\\
}

\newcommand{\yr}{\textcolor{black}}
\newcommand{\wry}{\textcolor{black}}

\begin{document}

\twocolumn[{%
\maketitle

\begin{center}
\vspace{-0.3in}
\includegraphics[width=0.97\textwidth]{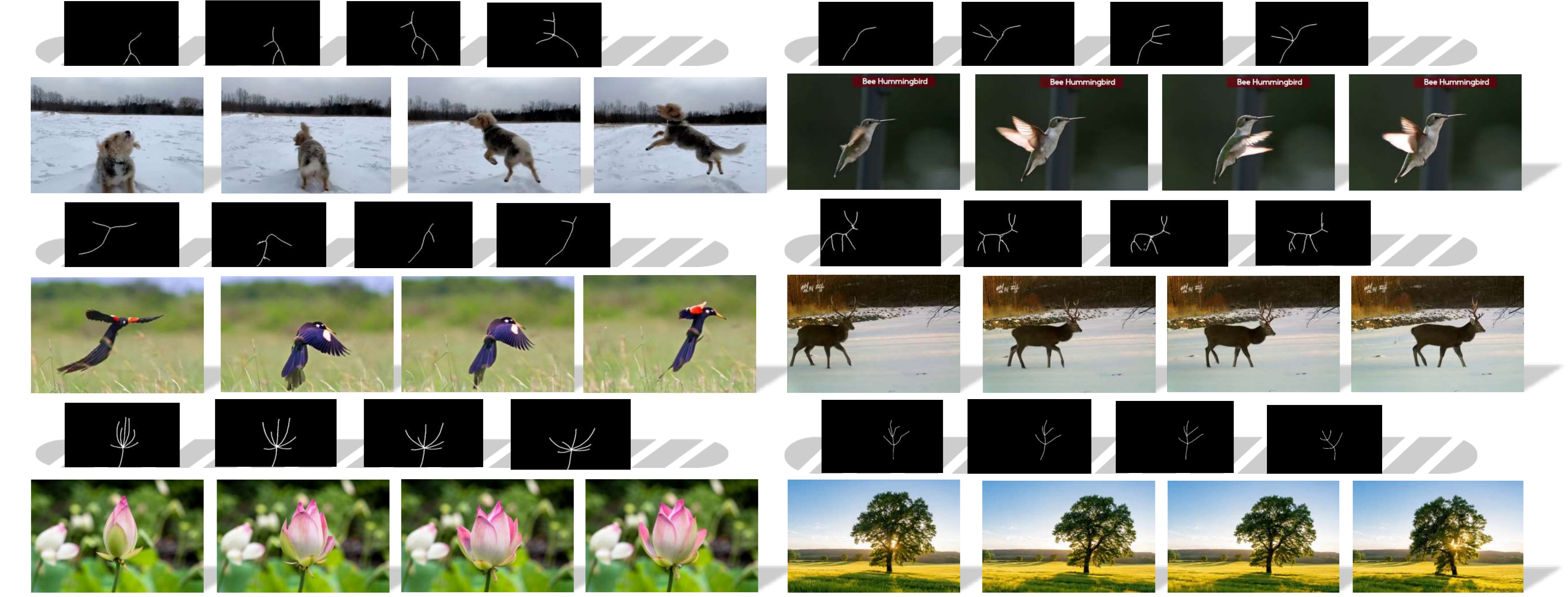}
\vspace{-0.05in}
\captionof{figure}{\textbf{PoseAnything}: The first universal pose-guided video generation model. It accepts a first frame and an arbitrary subject skeletal sequence as input to generate video clips conforming to the specified motion.}
\label{fig:teaser}
\vspace{-0.05in}
\end{center}
}]

\maketitle

{%
\renewcommand{\thefootnote}{}%
\footnotetext{\textsuperscript{*} Equal Contribution.}%
\footnotetext{\textsuperscript{\dag} Corresponding Author.}%
}

\begin{abstract}
Pose-guided video generation refers to controlling the motion of subjects in generated video through a sequence of poses. It enables precise control over subject motion and has important applications in animation. However, current pose-guided video generation methods are limited to accepting only human poses as input, thus generalizing poorly to pose of other subjects. To address this issue, we propose \textbf{PoseAnything}, the first universal pose-guided video generation framework capable of handling both human and non-human characters, supporting arbitrary skeletal inputs. To enhance consistency preservation during motion, we introduce Part-aware Temporal Coherence Module, which divides the subject into different parts, establishes part correspondences, and computes cross-attention between corresponding parts across frames to achieve fine-grained part-level consistency. Additionally, we propose Subject and Camera Motion Decoupled CFG, a novel guidance strategy that, for the first time, enables independent camera movement control in pose-guided video generation, by separately injecting subject and camera motion control information into the positive and negative anchors of CFG. Furthermore, we present \textbf{XPose}, a high-quality public dataset containing 50,000 non-human pose-video pairs, along with an automated pipeline for annotation and filtering. Extensive experiments demonstrate that Pose-Anything significantly outperforms state-of-the-art methods in both effectiveness and generalization.
\end{abstract}  
\section{Introduction}
\label{sec:intro}
Recent advancements in video generation~\cite{wan2025wanopenadvancedlargescale,kong2025hunyuanvideosystematicframeworklarge,hu2025high,hu2025hunyuancustom,chen2025instancev} have demonstrated remarkable capabilities in high-resolution synthesis~\cite{hu2025ultragen, xue2025ultravideo}, video editing~\cite{liang2025omniv2v,chen2025ivebench}, and diverse multi-modal tasks ranging from multi-subject interaction to audio-visual generation~\cite{hu2025polyvivid, hu2025hunyuancustom, hu2025harmony,zhang2025uniavgen}. Amidst these developments, precise motion controllability remains a critical pursuit, as explored in camera motion transfer~\cite{hu2024motionmaster} and trajectory controls~\cite{wang2024motionctrl}. Pose-guided video generation specifically addresses this by controlling the motion of subjects through a sequence of poses. By leveraging explicit pose information, it overcomes the limitations of traditional video generation methods, which often struggle to accurately and flexibly manipulate character poses and movements, thus demonstrating substantial potential for a wide range of applications, including entertainment video production, personalized animation, and performance-driven avatar animation.


The rapid advancement of diffusion models has led to numerous methods for pose-guided video generation based on this architecture. For example, Disco \cite{wang2024discodisentangledcontrolrealistic} modifies Stable Diffusion and incorporates background features via ControlNet, but struggles to preserve detailed character features and suffers from inter-frame jitter. AnimateAnyone~\cite{hu2024animateanyoneconsistentcontrollable} introduces ReferenceNe to better retain character appearance and improve pose control and temporal coherence, yet generating natural and continuous movements remains challenging. Although extensive research has been conducted in pose-guided video generation, they only focus on human-pose driven video generation.
Recent work like\wry{Animate-X~\cite{tan2024animatexuniversalcharacterimage} \yr{explores pose-guided video synthesis for non-human subjects by adapting human skeletons, but cannot accommodate diverse non-human skeletal structures. Currently, no method supports skeleton-driven video generation for arbitrary skeleton types.}}

\yr{Beyond pose-guided video generation, other} studies have adopted other 
conditions \yr{to control the motion in the generated video}.
\yr{These include}
trajectory-based controllable generation \yr{methods} (e.g., TORA~\cite{zhang2025toratrajectoryorienteddiffusiontransformer}, SG-I2V~\cite{namekata2025sgi2vselfguidedtrajectorycontrol}, ATI~\cite{wang2025atitrajectoryinstructioncontrollable} and LeviTor~\cite{wang2025levitor3dtrajectoryoriented}),
\yr{which} excel at guiding object 
position changes 
\yr{(\textit{e.g.,} overall translation, scaling),} but lack the 
\yr{granularity} to precisely control pose variations \yr{and part-level movements}. 
Sketch-based controllable generation \yr{methods} (e.g., SketchVideo~\cite{wang2025atitrajectoryinstructioncontrollable}), \yr{on the other hand, often require labor-intensive input and can struggle to maintain temporal consistency across frames,
highlighting the need for a more versatile and precise control mechanism.} 

\wry{To \yr{address these challenges,}
we propose \textbf{PoseAnything}, \yr{the first unified framework that} supports pose\yr{-guided video generation for} both human and non-human characters, \yr{accommodating \textbf{universal skeletal inputs}}. 
\yr{To address the limitations of} current methods in maintaining appearance consistency during motion, we introduce the \textbf{Part-aware Temporal Coherence Module}.
\yr{This module ensures fine-grained, part-level consistency by first partitioning the subject into distinct parts, establishing correspondences between them across frames, and then computing cross-attention exclusively among these matched parts,}
\yr{thus} refining the control granularity to the part level \yr{and guaranteeing superior temporal coherence.} 
Furthermore, we propose the \textbf{Subject and Camera Motion Decoupled CFG}, \yr{a novel guidance strategy that}, for the first time, \yr{enables} controllable camera movement in pose-guided video generation. 
By injecting subject and camera motion control information into the positive and negative anchors of CFG respectively, \yr{it} effectively \yr{decouples the two processes, resolving the} mutual interference \yr{that} occurs when 
both 
\yr{are injected in a coupled manner.}
Extensive quantitative and qualitative experiments validate the effectiveness of our model,
\yr{excelling in} preserving appearance consistency while allowing flexible control over both subject and camera movement. }
\wry{In addition, to support the universal pose-guided video generation task, we release \textbf{XPose}, the first public high-quality non-human pose dataset, \yr{consisting of 50,000 non-human pose-video pairs}. We design a pose extraction pipeline and a selection algorithm to extract precise and temporally continuous pose sequences from videos, providing a strong foundation for future research in related field.}

The main contributions of our work are three-fold:
\begin{itemize}
    \item \wry{We propose \textbf{PoseAnything}, the first unified framework that supports pose-guided video generation for both human and non-human characters\yr{, accommodating arbitrary skeletal inputs. In addition,} 
    we construct XPose, a high-quality public dataset \yr{comprising 50,000 non-human pose-video pairs}, laying a solid foundation for future research in this field.}

    \item \wry{We design a \textbf{Part-aware Temporal Coherence Module} \yr{to} address the challenge of maintaining subject consistency during motion. \yr{The module ensures fine-grained, part-level consistency by} part segmentation, establishing part correspondences, and part-aware cross-attention, \yr{thereby refining control granularity to} a finer level.}
    
    \item \wry{We propose \textbf{Subject and Camera Motion Decoupled CFG}, enabling camera motion control \yr{in} pose-guided video generation for the first time. \yr{It} effectively 
    \yr{decouples the subject and camera motion by separately injecting their control conditions into the positive and negative anchors of CFG, eliminating their mutual interference.}
    }
    
\end{itemize}


\section{Related Works}
\label{sec:related_work}
\begin{figure*}[htbp]
    \centering
    \includegraphics[width=0.9\linewidth]{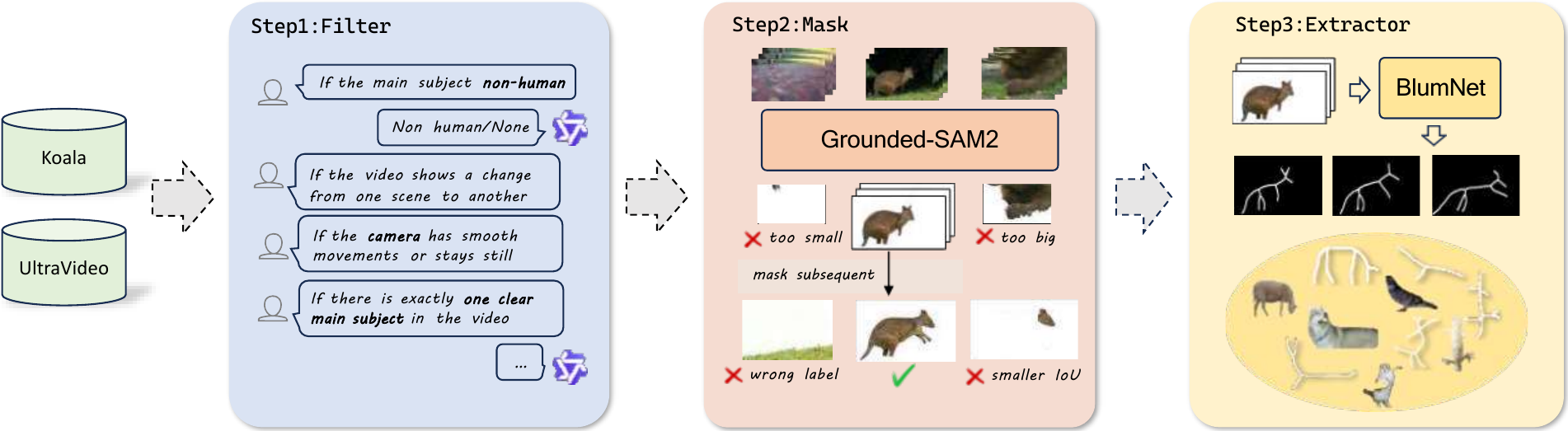}
    \caption{Construction pipeline of \textbf{XPose} \yr{dataset}. This process consists of three stages: (1) selecting videos with \textbf{non-human} subjects from the Koala and UltraVideo datasets; (2) employing Grounded-SAM2 to segment the subjects in the videos and applying a \textbf{filtering algorithm} to select high-quality mask sequences; and (3) \textbf{extracting} poses of the masked subjects using BlumNet.}
    \vspace{-0.1in}
    \label{fig:data-pipeline}
\end{figure*}
\subsection{Diffusion Models for Video Generation}
In recent years, diffusion models have achieved rapid development. Early methods, such as Stable Video diffusion \cite{blattmann2023stablevideodiffusionscaling}, primarily used U-Net-based architectures with 3D convolutional layers for temporal modeling. Following the release of Sora \cite{liu2024sorareviewbackgroundtechnology}, DiT-based approaches \cite{peebles2023scalablediffusionmodelstransformers}) have increasingly replaced UNet-based methods in the field, capitalizing on the superior scalability and context-awareness of the Transformer architecture. Other DiT-based models like HunyuanVideo \cite{kong2025hunyuanvideosystematicframeworklarge} Wan \cite{wan2025wanopenadvancedlargescale} MovieGen \cite{polyak2025moviegencastmedia} utilize a 3D causal VAE \cite{yu2023magvitmaskedgenerativevideo} to handle the encoding and decoding of raw video data. \yr{Since} these models acquire knowledge of inter-frame consistency and temporal continuity during pretraining, their application to character animation tasks enhances realism in generated characters and improves temporal coherence by implicitly capturing long-range dependencies within the video sequence. Our Pose-Anything framework is built upon the open-source model Wan2.2-TI2V-5B \cite{wan2025wanopenadvancedlargescale}, leveraging its robust pretrained capabilities to ensure high-quality visual generation from the outset.

\subsection{Pose-guided Video Generation}
Recent advances in human pose extraction, such as DWPose \cite{yang2023effectivewholebodyposeestimation} and DensePose \cite{güler2018denseposedensehumanpose}, enable human pose-guided video generation. 
Early works like Disco \cite{wang2024discodisentangledcontrolrealistic} focus on dance actions but were limited in terms of generalization, appearance consistency, and inter-frame continuity. Following this, AnimateAnyone~\cite{hu2024animateanyoneconsistentcontrollable} incorporates the ReferenceNet architecture to integrate character appearance features, thereby achieving remarkable performance in consistency preservation. Additionally, temporal layers are employed to facilitate temporal modeling.

Although many studies have been conducted in pose-guided video generation, they focus on human-centric video synthesis. Some works like Animate-X \cite{tan2024animatexuniversalcharacterimage} have attempted to generate videos with non-human characters. However, their driving factors are still limited to human \yr{skeletons} and movements. To the best of our knowledge, our Pose-Anything is the first controllable video generation model to enable pose-driven \yr{video} synthesis for arbitrary subjects.

\subsection{Universal Animation}
\begin{figure*}[htbp]
    \centering
    \includegraphics[width=0.9\linewidth]{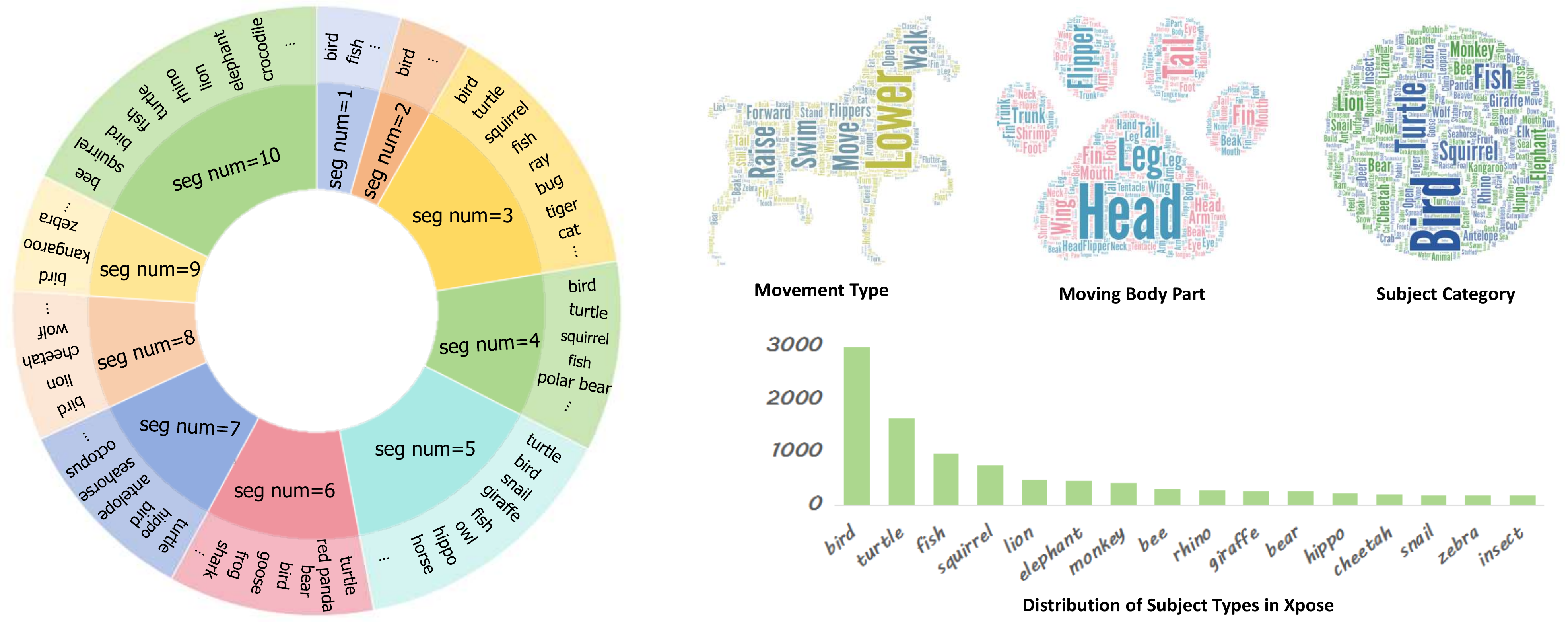}
    \caption{Comprehensive statistics of XPose in several aspects. \wry{The left shows the distribution of poses with different \textbf{numbers of segments} in XPose, as well as the corresponding \textbf{subject categories} for each segment count. The right illustrates the dataset distribution across three dimensions: \textbf{motion type}, \textbf{motion body part}, and \textbf{subject category}. As shown in the figures, our dataset exhibits good diversity.}}
    \vspace{-0.1in}
    \label{fig:dataset analysis}
\end{figure*}
Various methods have been proposed to inject control information for animation. Sketch-based approaches like VideoComposer~\cite{wang2023videocomposercompositionalvideosynthesis} and SketchVideo~\cite{liu2025sketchvideosketchbasedvideogeneration},  use pose sequences for motion guidance but are challenging to operate in practice. Beginning with DragNUWA \cite{yin2023dragnuwafinegrainedcontrolvideo}, some studies have explored motion control based on trajectory conditions, such as TORA \cite{zhang2025toratrajectoryorienteddiffusiontransformer}, SG-I2V \cite{namekata2025sgi2vselfguidedtrajectorycontrol}, ATI \cite{wang2025atitrajectoryinstructioncontrollable}, and LeviTor \cite{wang2025levitor3dtrajectoryoriented}. Nevertheless, these approaches struggle to capture fine-grained variations in subject pose. In contrast, our universal pose-guided video generation model, PoseAnything, not only enables flexible control over object positions but also provides precise manipulation of diverse poses.

\section{XPose: A Universal Pose Dataset}
\label{sec: XPose dataset}
\wry{Pose-driven video generation for arbitrary characters requires that includes a wide variety of subjects, however, such data are absent from existing public datasets. To address this gap, we release \textbf{XPose}, a high-quality public dataset comprising 50,000 \textbf{non-human} pose-video pairs. As the first dataset focused on non-human poses, XPose provides crucial data for pose-guided video generation with diverse entities, paving the way for future research and applications with a wider range of characters.}
\yr{As shown in} Fig. \ref{fig:data-pipeline}, \yr{our} construction pipeline \yr{consists of} three stages: video filtering, subject masking, and pose extraction.

\noindent\textbf{Stage 1: \yr{Video} Filter\yr{ing}.} To reduce noise in extracted skeletons, XPose focuses on videos featuring single non-human object. To filter out videos  that fail to satisfy the criteria, we employ Qwen-2.5-VL-7B-Instruct~\cite{qwen2.5-VL} to filter samples from Koala~\cite{wang2024koala36mlargescalevideodataset} and UltraVideo~\cite{xue2025ultravideohighqualityuhdvideo} datasets. 

\noindent\textbf{Stage 2: \yr{Subject} Mask\yr{ing}.} We utilize Grounded-SAM-2~\cite{ravi2024sam2segmentimages}to generate segmentation masks for the primary object in each video. At this stage, we design an algorithm to filter out invalid skeletons \yr{ensuring the consistency of the extracted subject across frames}. 
First, to ensure the completeness of the subject and valid motion information, we discard videos in which the mask region is excessively large or small. Specifically, we calculate the area $S$ of the masked subject and retain videos where the ratio of $S$ to the entire image \yr{$\frac{S_t}{H \times W}$} falls within the interval $(0.2,0.8)$. 
Second, as the Grounded-SAM-2 model produces multiple object masks, we select the largest mask in the first frame and designate its corresponding subject as the target \yr{:}
\begin{equation}
M_{1}^* = \arg\max_{M \in \mathcal{M}_1} \mathrm{Area}(M).
\end{equation}

For subsequent frames, we select masks with the same label as the first frame. If multiple candidates exist, we choose the one with the highest intersection-over-union (IoU) with the last selected mask to maintain temporal consistency:
\begin{equation}
M_t^* = \arg\max_{M \in \mathcal{M}_t, \mathrm{label}(M) = \mathrm{label}(M_1^*)} \mathrm{IoU}(M, M_{t-1}^*).
\end{equation}

If no valid mask is found, the frame is skipped.

\noindent\textbf{Stage 3: \yr{Pose} Extraction.} Finally, we apply BlumNet \cite{Zhang2022BlumNet} to extract skeletons from the masked images. If the number of frames with successfully extracted skeletons $T_{skel}$ in a video is less than $80\%$ of the total frames $T$, \yr{\textit{i.e.}, $\frac{T_{skel}}{T} < 0.8$,} the video is discarded. This data extraction pipeline and filtering algorithm effectively ensure the accuracy of the extracted skeleton information and the temporal consistency across frames, thereby laying a solid foundation for the construction of high-quality datasets.

\noindent\wry{\textbf{Dataset Analysis.} Fig.~\ref{fig:dataset analysis} presents a comprehensive statistical overview of XPose. 
The left panel shows the distribution of skeleton segment counts and corresponding subject categories. The word cloud illustrates the diversity and frequency of subject types, motion types, and motion parts, while the bar chart presents the distribution of subject categories. Together, these analyses demonstrate the richness and diversity of the XPose dataset.}

\section{Method}
\label{sec:method}
\begin{figure*}[htbp]
    \centering
    \includegraphics[width=0.95\linewidth]{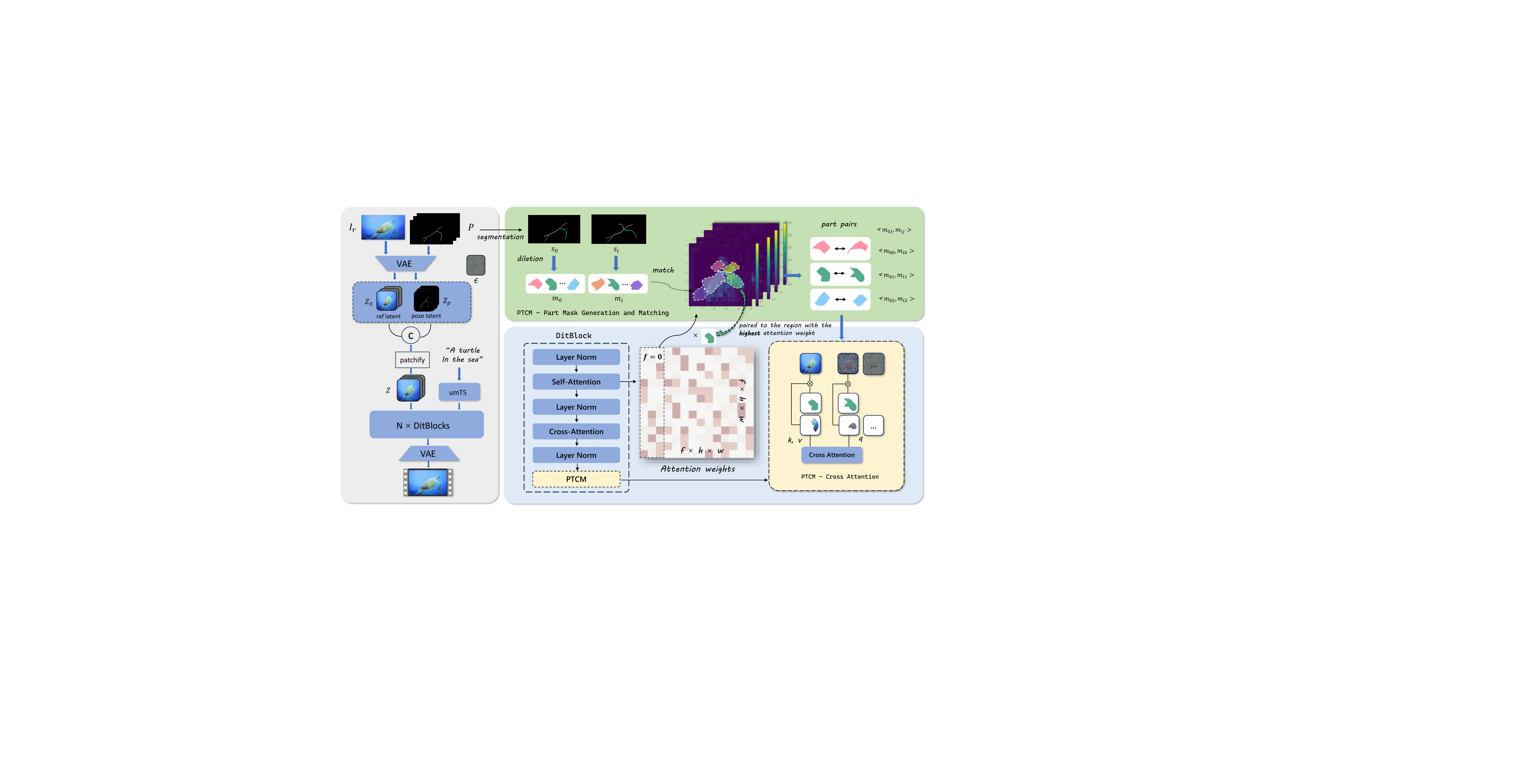}
    \caption{Overview of our \textbf{PoseAnything}. 
    \wry{Given a reference image $I_r$ and a pose sequence $P$, we first encode $P$ into \yr{pose} latent $Z_p$, and then concatenate it with the latent $Z_0$ of $I_r$ along the channel dimension. Additionally, \yr{we propose} \textbf{Part-aware Temporal Coherence Module} \yr{for fine-grained appearance consistency control}: 1) We segment the pose into separate segments $s_{ij}$ and dilate each segment to obtain the subject part masks $m_{ij}$; 2) We then use attention patterns to match the same parts across different frames; 3) For each pair $<m_{0j}, m_{ij}>$, we introduce a part-aware cross-attention module in the DiTBlock to compute cross-attention between \yr{matched parts}. 
    By performing consistency control at the part level, \yr{the} part-aware coherence module \yr{achieves enhanced} subject appearance consistency in a fine-grained manner.}
 }
 \vspace{-0.1in}
    \label{fig:framework}
\end{figure*}
\wry{Pose-guided video generation takes a sequence of poses as input, alongside a reference image and a textual prompt\yr{, and aims to generate a video} whose subject movement faithfully aligns with the specified pose sequence. 
In contrast to previous studies, our proposed \textbf{PoseAnything} is capable of accommodating \textbf{universal skeletal inputs}, \yr{including both non-human and human poses}, which is the first work to accomplish this task. 
\yr{Moreover}, we introduce \yr{\textbf{Part-aware Temporal Coherence Module}}, a fine-grained mechanism for controlling appearance consistency \yr{across frames}. This method involves partitioning the subject into multiple parts\yr{, establishing part correspondence} and computing cross attention 
\yr{between matched parts} across frames, thereby facilitating 
\yr{enhanced finer-grained part-level consistency control.}
Furthermore, we develop \yr{\textbf{Subject and Camera Motion
Decoupled CFG},} a CFG-based decoupled control method for subjects and camera movements. \yr{It} separately injects subject pose control conditions and camera motion control conditions into the positive and negative \yr{anchors}, respectively, effectively mitigating mutual interference between the two types of motion conditions.}

The overall framework \yr{of PoseAnything} is illustrated in Fig.~\ref{fig:framework}, 
\yr{based on} Wan2.2-TI2V-5B~\cite{wan2025wanopenadvancedlargescale}. 
\wry{We fuse the latent representation of \yr{the} reference image and pose by concatenating them along the channel dimension as the input of DiTBlock. The \yr{Part-aware Temporal Coherence Module} is incorporated after the original cross-attention layer within each DiTBlock, with the aim of enhancing appearance consistency at a finer-grained level} 
\yr{(detailed} in Sec.~\ref{subsec:Instance-Wise Cross Attention}).

\subsection{\wry{Analysis of Condition Injection Strategies}}
\label{sec: condition injection strategies}
We utilize Wan2.2-TI2V-5B \cite{wan2025wanopenadvancedlargescale} as our base \yr{model for} image to video \yr{generation}. 
As shown in Fig.~\ref{fig:framework}, the original model takes a reference image $I_r$ and employs the pretrained Wan2.2VAE to encode it into a latent representation $Z_i$. 
$Z_i$ \yr{is} concatenated with noise latent $\epsilon$ along the temporal dimension and patchified to form the input $Z_0$ of the DiTBlock. 
\yr{For pose-guided generation, an} additional pose sequence $P$ \yr{is taken} as input. 
For easy integration, we encode \yr{the} pose sequence $P$ into \yr{pose} latent presentations $Z_p$, \yr{also using the} pretrained Wan2.2 VAE.
To effectively incorporate skeletal information, we compare three different injection strategies: concatenation by channels, multi-layer perceptron (MLP), and concatenation by width.\\
\textbf{Strategy 1: Concatenation by channel.} Given the initial latent $Z_i$ and the pose latent $Z_p$, we first concatenate $Z_i$ with a noise map $\epsilon$ to get $Z_0=[Z_i, \epsilon]$. Next, $Z_0$ and $Z_p$ are concatenated along the channel axis to obtain the aggregated latent $Z_{agr}$. In the patchify module, we increase the number of input channels for the convolutional layers to accommodate the additional skeleton dimensions, while maintaining the channel number of input of \yr{DiT block} consistent with \yr{the} original Wan model:
\begin{equation}
\begin{aligned}
&Z_{agr} = [Z_0, Z_p]  \in F \times H \times W \times 2C,\\
&Z = Conv(Z_{agr}) \in f \times h \times w \times c. 
\end{aligned}
\end{equation}
\textbf{Strategy 2: Multi-layer Perceptron.} $Z_p$ is converted to the same shape as \wry{$Z_0$} using a MLP. The resulting features are then fused with the initial latent \wry{$Z_0$} by element-wise addition, yielding the DiT model input $Z$:
\begin{equation}
Z = {Z_0} + \mathrm{MLP}(X_p).
\end{equation}
\textbf{Strategy 3: Concatenation by width.} \wry{$Z_0$} and $Z_p$ are concatenated along the width dimension to form an aggregated latent $Z$, which is directly used as the input of DiT block:
\begin{equation}
\begin{aligned}
&Z = \mathrm{Concat}_\mathrm{width}(Z_0, Z_p) \in F \times H \times 2W \times C.
\end{aligned}
\end{equation}
\textbf{\yr{Comparison of Injection Strategies}.} Our experimental results demonstrate that channel-wise conditioning methods exhibit significant advantages in pose-guided video generation (presented in \#Suppl). Consequently, we employ this strategy to inject skeletal information into our model, enabling more effective utilization of poses.

\subsection{\yr{Part-aware Temporal Coherence Module}}
\label{subsec:Instance-Wise Cross Attention}
\wry{
\yr{In pose-guided video generation, existing methods often struggle}
to maintain the consistency of the object’s appearance throughout the motion\yr{, especially during large pose changes}. 
Previous works attempt to address this issue by utilizing ControlNet or cross-attention mechanisms to capture the overall appearance of the object in the reference image. However, these approaches still face inconsistencies or distortions in fine details. To tackle this, we propose a \textbf{Part-aware Temporal Coherence Module (PTCM)}. 
We divide the object into multiple smaller parts, utilize attention weight to match corresponding parts across different frames, and perform cross-attention between the matched parts. By decomposing overall appearance consistency into \textit{finer-grained part-level consistency control}, our method achieves superior performance in maintaining temporal coherence.} \yr{As shown in Fig.~\ref{fig:framework}, the Part-aware Temporal Coherence Module (PTCM) consists of three steps.}

\noindent\textbf{Part Mask Generation.} We first segment each pose into segments, denoted as $s_{ij}$, where $i$ denotes the frame index corresponding to the skeleton image, and $j$ denotes the index of the segment of \yr{the} current pose. To obtain the pixels $m_{ij}$ corresponding to $s_{ij}$, we dilate each $s_{ij}$ by $\alpha$:
\begin{equation}
m_{ij} = \mathrm{Dilate}(s_{ij}, \alpha),
\end{equation}
\yr{where the expansion coefficient} $\alpha$ \yr{is calculated by} continuously 
\yr{dilating the skeleton}
until it can cover the main body in the reference image:
\begin{equation}
\begin{small}
\alpha_{ij} = \min \left\{ \alpha, 100 \;\middle|\; \mathrm{IoU}\left( \mathrm{Dilate}(s_{ij}, \alpha), \mathrm{Body}_{ij}^{\mathrm{ref}} \right) \geq 1 \right\}.
\end{small}
\end{equation}

\noindent\textbf{Part Matching using Attention Patterns.} 
\yr{Next, we establish correspondences between parts across frames.}
\yr{Based on} the observation that the attention weight between the same parts in different frames is higher than that between different parts,
\wry{we match each part in the first frame to its counterpart in subsequent frames by}:
\begin{equation}
s_{ij'} \thicksim s_{0j}\iff j'=\mathop{\arg\max}\limits_{t} \ \text{attn\_weight}[m_{0j}][m_{it}].
\end{equation}

\wry{In implementation, we first perform several steps of inference to compute the attention weights between the first frame and subsequent frames. Then, using the aforementioned method, we match the masks of the first frame to those of the subsequent frames based on these attention weights, as shown in Fig.~\ref{fig:framework}.}

\noindent\textbf{Part-aware Cross Attention.} 
\begin{figure}[t]
    \centering
    \includegraphics[width=0.75\linewidth]{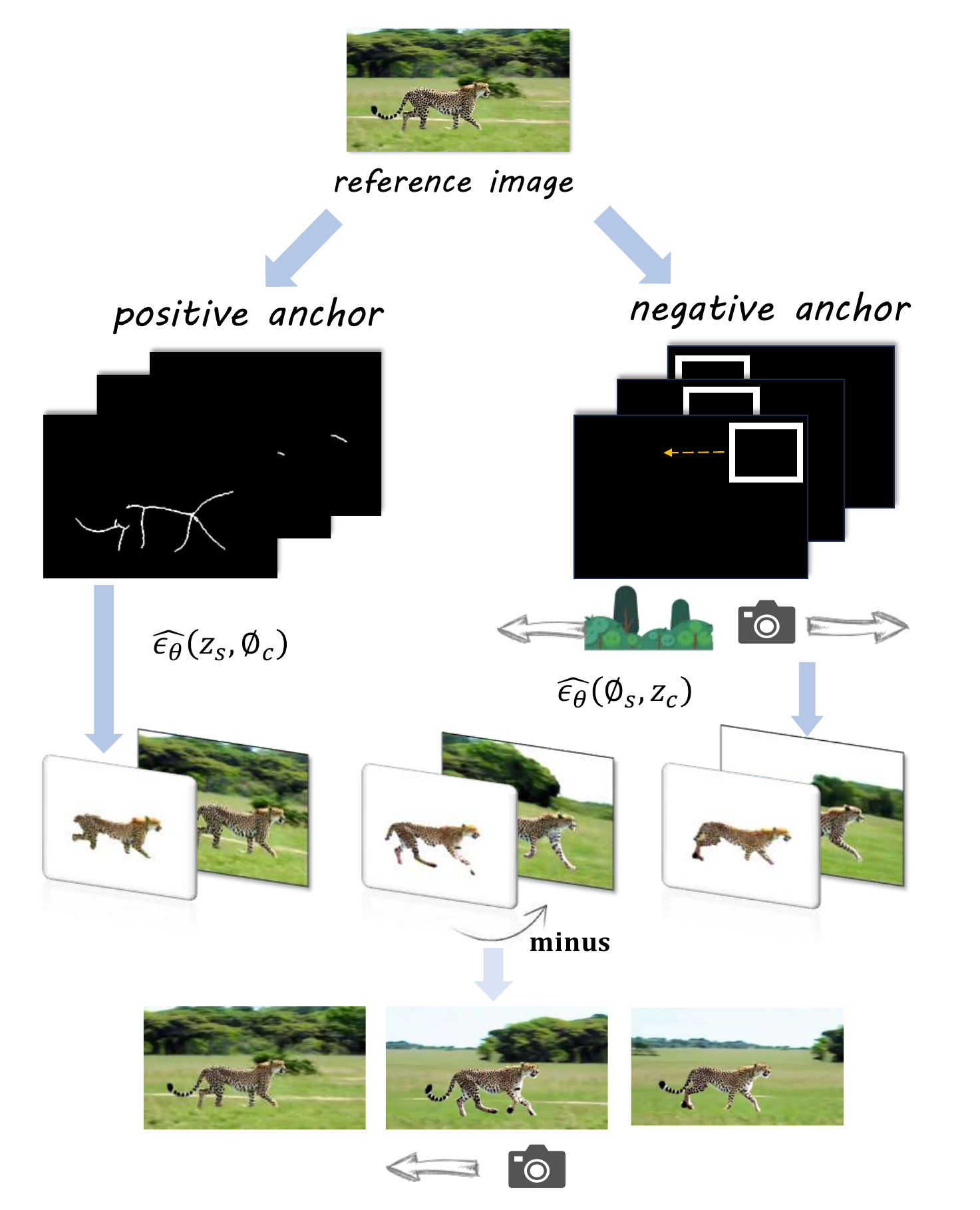}
    \vspace{-0.1in}
    \caption{Subject and Camera Decoupled Control Based on CFG.}
    \vspace{-0.1in}
    \label{fig:motion control}
\end{figure}
For \yr{each} pair $<s_{0j}, s_{ij}>$, we calculate cross-attention by calculating \yr{K} and \yr{V} using the tokens corresponding to \yr{$s_{0j}$} in the first frame, and calculate \yr{Q} using the tokens corresponding to $s_{ij}$ in subsequent frames:
\begin{equation}
\begin{aligned}
&x\yr{'} = x + \text{Cross-Atten}(Q_j, K_j, V_j),\\
&Q_j = m_{ij}XW_q, K_j = m_{0j}X_0W_k, V_j = m_{0j}X_0W_v.
\end{aligned}
\end{equation}

\wry{This module is inserted after the final cross-attention layer in the DiT block, as shown in Fig.~\ref{fig:framework}}.

\subsection{\wry{Subject and Camera Motion Decoupled CFG}}

Current pose-guided video generation methods are limited to controlling object motion and do not support background/camera motion control. Other video generation approaches, such as SG-I2V (based on drag), inject both object and camera motion control conditions simultaneously. However, such a coupled injection strategy often results in mutual interference between the two control conditions, hindering the model’s ability to comprehensively represent both types of motion information.To address this critical limitation, we ingeniously leverage the positive and negative anchors of Classifier-Free Guidance (CFG) to decouple subject and camera motion control (Fig.~\ref{fig:motion control}). This innovative approach enables a complete separation of the two control conditions and effectively prevents mutual interference. As a result of this decoupling design, the model can learn and precisely execute the respective motion instructions more clearly, significantly enhancing the motion fidelity and controllability of the generated videos.

\noindent{\textbf{\yr{Decoupled Subject and Camera Motion via CFG.}}}
\begin{figure*}[htbp]
    \centering
    \includegraphics[width=0.96\linewidth]{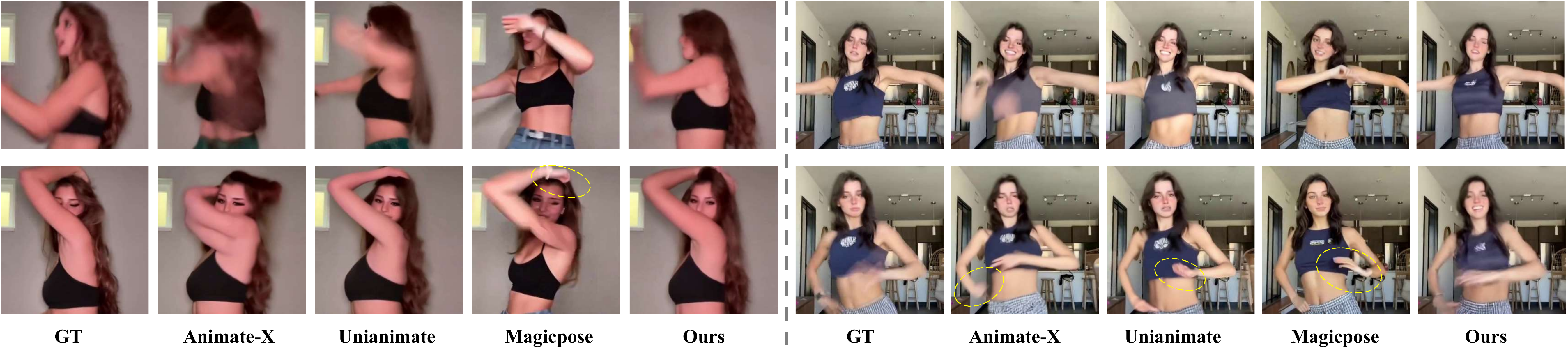}
    \caption{Quantitative comparison between the state-of-the-arts and
 Ours on TikTok dataset. }
    \label{fig:human-comparison}
\end{figure*}
\wry{In practice, we find that although our model is trained on subject motion control, it can be generalized to control camera motion as well. However, directly injecting both subject and camera motion control conditions simultaneously leads to mutual interference between the two control signals. To tackle this, we propose decoupled subject and camera motion control via classifier-free guidance: 
\yr{injecting the subject motion control conditions (pose sequence) into the positive anchors of CFG, while injecting the camera motion control conditions into the negative anchors.}
The underlying principle is illustrated as follows:}
\vspace{-0.05in}
\begin{equation}
\begin{aligned}
&\tilde{\epsilon}=\hat{\epsilon}_{\theta}(\emptyset_{s},z_{c})+s\cdot\left(\hat{\epsilon}_{\theta}(z_{s},\emptyset_{c})-\hat{\epsilon}_{\theta}(\emptyset_{s},z_{c})\right)\\
&=\hat{\epsilon}_{\theta}(z_{s},\emptyset_{c})-\hat{\epsilon}_{\theta}(\emptyset_{s},\emptyset_{c})-s\cdot[\hat{\epsilon}_{\theta}(\emptyset_{s},z_{c})-\hat{\epsilon}_{\theta}(\emptyset_{s},\emptyset_{c})]\\
&=(1+s)\cdot\hat{\epsilon}_{\theta}(\emptyset_{s},\emptyset_{c})+\hat{\epsilon}_{\theta}(zt_{s},\emptyset_{c})+s\cdot\hat{\epsilon}_{\theta}(\emptyset_{s},zt_{c}),
\end{aligned}
\end{equation}
where $Z_c$ denotes the latents injected with camera motion information, and $Z_s$ denotes the latents injected with subject motion information.\\
\noindent\textbf{\yr{Camera Control via Negative Anchors.}}
\yr{Our key idea is to use the negative anchor in CFG to steer the generation away from specific camera states, thereby achieving camera movement control.
This requires the control signal injected into the negative anchor to be \textit{opposite} to be the target camera motion.}
\wry{
\yr{For instance, to generate a leftward camera movement (where the background should move rightward), we generate a skeleton sequence with a rectangle that moves \textit{leftward} and inject it into the negative anchor as shown in Fig.~\ref{fig:motion control}.}
\yr{Injecting this ``left-moving" negative signal prompts the model to produce a rightward background flow -- achieving the desired leftward camera pan.} 
\yr{This decoupled CFG design effectively enables} precise and independent control over both the subject and the camera motion.}

\section{Experiments}
\subsection{\yr{Experiment} Settings}
 \textbf{Implementation Details.} We utilize the XPose dataset alongside 15,000 internal human videos as train\yr{ing} set. The training process is divided into three stages. In the first stage, we train a baseline model without the part-aware temporal coherence module solely on \yr{the} human dataset for 3k iterations, with batch size set to 32 and learning rate 5e-5. In the second stage, we further trained the above model with human \yr{and} non-human mixed data with the same batch size and learning rate. In the third stage, we exclusively train the part-aware temporal coherence module on the mixed dataset while keeping all other modules frozen for 8k iterations, with batch size set to 32 and learning rate 1e-5. All experiments are performed on 8 GPUs. \\
\wry{\textbf{Evaluation Details.} To comprehensively evaluate the model’s performance on both human and non-human data, we conduct qualitative and quantitative experiments separately. The generated videos are assessed using five standard metrics: (1) PSNR, (2) SSIM, (3) L1 distance, (4) LPIPS, and (5) FVD.}
\begin{figure*}[h]
    \centering
    \includegraphics[width=0.99\linewidth]{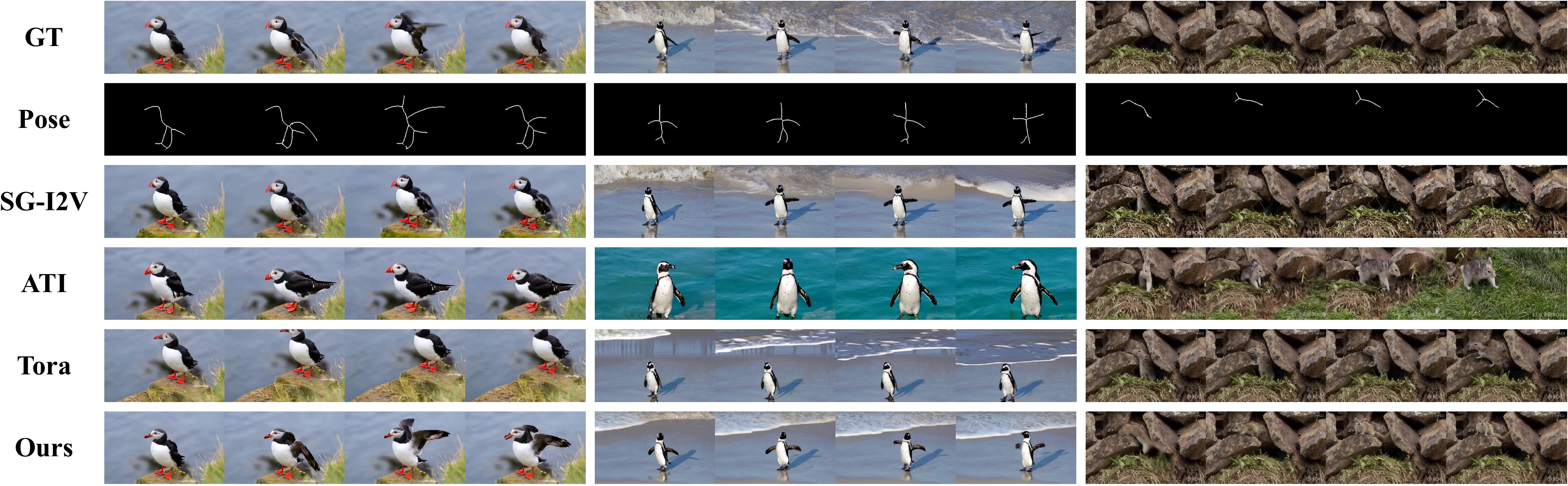}
    \vspace{-0.1in}
    \caption{ Qualitative comparison with existing state-of-the-art methods on XPose-benchmark. }
    \vspace{-0.1in}
    \label{fig:non-human-comparison}
\end{figure*}

\subsection{Human Pose-Guided Generation }
\wry{To validate the effectiveness of our method on human data, we conduct experiments on the widely-used benchmark, TikTok~\cite{jafarian2022selfsupervised3drepresentationlearning}. To ensure a fair comparison, we separately train our Pose Anything for 1,500 iterations exclusively on the training set of the TikTok dataset. Both qualitative and quantitative experiments are then conducted on the test split.} We compare our model with several state-of-the-art methods, including Disco \cite{wang2024discodisentangledcontrolrealistic}, MagicAnimate \cite{xu2023magicanimatetemporallyconsistenthuman}, Animate Anyone \cite{hu2024animateanyoneconsistentcontrollable}, Champ \cite{zhu2024champcontrollableconsistenthuman}, Unianimate \cite{wang2024unianimatetamingunifiedvideo}, Animate-X \cite{tan2024animatexuniversalcharacterimage}.
\noindent\textbf{Quantitative results.} The quantitative comparison results between our method and the state-of-the-arts on Tiktok are reported in Tab.~\ref{tab:human-comparison}. PoseAnything achieves the best performance across all metrics, showcasing its superior ability to generate highly controllable and photorealistic human videos.
\noindent\textbf{Qualitative results.} Fig.~\ref{fig:human-comparison} shows qualitative comparison results of our approach with UniAnimate, MagicPose, and Animate-X. It can be observed that while the results generated by other methods exhibit obvious distortions, PoseAnything demonstrates excellent motion alignment and appearance consistency. 
\begin{table}[t]
    \centering
     \small
      \vspace{-0.1in}
    \caption{Quantitative comparisons with the state-of-the-arts on TikTok dataset \yr{(Human)}.}
    \vspace{-0.1in}
     \resizebox{\columnwidth}{!}{
    \begin{tabular}{lccccc} 
        \toprule
        Method & PSNR$\uparrow$ & SSIM$\uparrow$ & L1$\downarrow$ & LPIPS$\downarrow$ & FVD$\downarrow$ \\
        \midrule
        Disco~\cite{wang2024discodisentangledcontrolrealistic}             & 29.03 & 0.668 & 3.78E-04 & 0.292 & 292.80 \\
        MagicAnimate~\cite{xu2023magicanimatetemporallyconsistenthuman}      & 29.16 & 0.714 & 3.13E-04 & 0.239 & 179.07 \\ 
        MagicPose~\cite{chang2023magicpose}             & 29.53 & 0.752 & 0.81E-04 & 0.292 & - \\
        AnimateAnyone~\cite{hu2024animateanyoneconsistentcontrollable}       & 29.56 & 0.718 & - & 0.285 & 171.90 \\
        Champ~\cite{zhu2024champcontrollableconsistenthuman}       & 29.91 & 0.802 & 2.94E-04 & 0.234 & 160.82 \\
        Unianimate~\cite{wang2024unianimatetamingunifiedvideo}       & 30.77 & 0.811 & 2.66E-04 & 0.231 & 148.06 \\
        Animate-X~\cite{tan2024animatexuniversalcharacterimage}      & 30.78 & 0.806 & 2.70E-04 & 0.232 & 139.01 \\
         \rowcolor{gray!10}
        \textbf{PoseAnything}       & \textbf{31.50} & \textbf{0.836} & \textbf{2.79E-05} & \textbf{0.224} & \textbf{133.95} \\
        \bottomrule
  \end{tabular}}
    \label{tab:human-comparison}
\end{table}

\subsection{Non-human Pose-guided Generation}
\begin{figure}[h]
    \centering
    \includegraphics[width=0.95\linewidth]{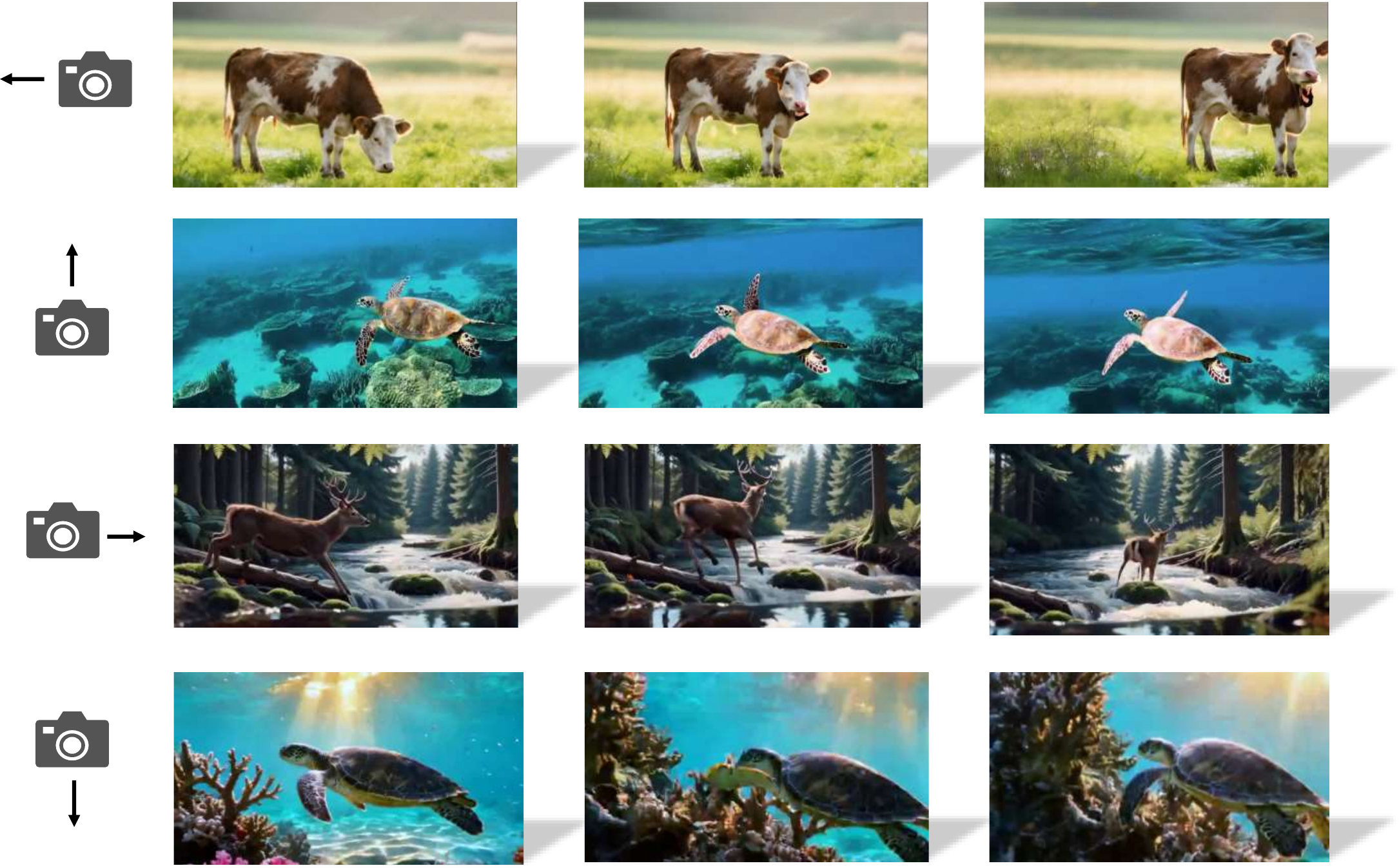}
    \vspace{-0.1in}
    \caption{Demonstration of Camera Control Cases.}
    \vspace{-0.1in}
    \label{fig:camera control demo}
\end{figure}

As there is no existing universal pose-guided video generation, we compare our method with controllable generation approaches based on drag-and-control methods, including ATI~\cite{wang2025atitrajectoryinstructioncontrollable}, Tora~\cite{zhang2025toratrajectoryorienteddiffusiontransformer}, and SG-I2V~\cite{namekata2025sgi2vselfguidedtrajectorycontrol}. We conduct comparison experiments on 51 videos randomly selected from XPose. For a fair comparison, these videos were not used during training.
\begin{table}[t]
    \centering
     \small
     \setlength{\tabcolsep}{4pt}
    \caption{Quantitative comparison between the state-of-the-arts and Ours on XPose-benchmark \yr{(Non-human)}.}
\vspace{-0.1in}
    \begin{tabular}{lccccc} 
        \toprule
        Method & PSNR$\uparrow$ & SSIM$\uparrow$ & L1$\downarrow$ & LPIPS$\downarrow$ & FVD$\downarrow$ \\
        \midrule
        Tora~\cite{zhang2025toratrajectoryorienteddiffusiontransformer}     & 30.08 & 0.6929 & 9.38E-06 & 0.3530 & 103.75 \\
        ATI~\cite{wang2025atitrajectoryinstructioncontrollable}      & 30.15 & 0.6810 & 9.59E-06 & 0.3706 & 101.44 \\ 
        SG-I2V~\cite{namekata2025sgi2vselfguidedtrajectorycontrol}       & 29.86 & 0.6634 & 1.28E-05 & 0.3674 & 102.97 \\
        \rowcolor{gray!10}
        \textbf{PoseAnything}       & \textbf{30.29} & \textbf{0.7114} & \textbf{8.19E-06} & \textbf{0.3241} & \textbf{99.97} \\
        \bottomrule
  \end{tabular}
  \vspace{-0.2in}
    \label{tab:non-human-comparison}
\end{table}
\noindent\textbf{Quantitative results} are shown in Table~\ref{tab:non-human-comparison}, demonstrating that our PoseAnything achieves the best performance \yr{in non-human pose-guided generation}. 
\noindent\textbf{Qualitative results} are presented in Fig.~\ref{fig:non-human-comparison}, from which we can observe that ATI, SG-I2V, and Tora fail to achieve accurate object pose control. Furthermore, these approaches often result in object deformation and background artifacts handling large-magnitude motions. In contrast, our PoseAnything model generates accurate object motions based on skeletal guidance, while simultaneously preserving the integrity and realism of both the object and the background. 

\subsection{Camera Motion Control}
To validate the effectiveness and robustness of our \textbf{Subject and Camera Motion Decoupled CFG}, we designed a challenging set of experiments. In this setup, we task the model with handling two distinct and concurrent motion signals: the subject is driven by its own dynamic pose sequence, while a separate camera motion command (e.g., pan left, tilt up) is simultaneously injected into the negative anchor of the CFG. This scenario directly tests our core claim of preventing mutual interference between subject and camera controls. The qualitative results, presented in Fig.~\ref{fig:camera control demo}, showcase the remarkable success of our approach. The subject accurately performs its intended actions according to the pose guidance, while the camera simultaneously executes the specified movement smoothly and coherently. The ability to maintain high fidelity for both the subject's action and the global camera motion provides strong empirical evidence that our method effectively disentangles the two control signals, achieving the precise and independent manipulation it was designed for.

\subsection{Ablation Study}
\textbf{Ablation on Part-aware Temporal Coherence Module.}
To assess the contribution of the Part-aware Temporal Coherence Module (PTCM), we conducted \yr{an ablation} study on XPose. Specifically, we compared the baseline model, which employs only concatenation for pose injection, with the full model integrating the PTCM module. Furthermore, we evaluated a configuration in which cross-attention is computed over the entire object region (EC) without part segmentation and matching. Quantitative results are summarized in \yr{Tab.~}\ref{tab:ablation}. \wry{Results show that the model without the PTCM module has poorer performance. Furthermore, omitting part segmentation and matching also leads to a degradation in model performance, which effectively \yr{validates} the contribution of the PTCM module.}
\begin{table}[t]
    \centering
    \small
    \setlength{\tabcolsep}{4pt}
        \caption{Quantitative results of ablation study.}
        \vspace{-0.1in}
    \begin{tabular}{lccccc} 
        \toprule
        Method & PSNR$\uparrow$ & SSIM$\uparrow$ & LPIPS$\downarrow$ & L1$\downarrow$ & FVD$\downarrow$ \\
        \midrule
        Concat   & 29.85 & 0.6964 & 0.3304 & 9.43E-06 & 102.30 \\ 
        EC       & 30.27 & 0.7107 & 0.3243 & \textbf{8.15E-06} & 101.50 \\ 
        PTCM      & \textbf{30.29} & \textbf{0.7114} & \textbf{0.3241} & 8.19E-06 & \textbf{99.970} \\ 
        \bottomrule
  \end{tabular}
    \vspace{-0.2in}
    \label{tab:ablation}
\end{table}

\section{Conclusion}
In this work, we introduce PoseAnything, the first unified framework supporting arbitrary skeletal inputs for pose-guided video generation. To address the challenge of maintaining consistent object appearance throughout motion sequences, we propose a Part-aware Temporal Coherence Module that enables fine-grained appearance consistency control at the part level. We are also the first to incorporate camera control by decoupling subject and camera motions through separate conditioning branches in classifier-free guidance, which enables a complete separation of the two control conditions and effectively prevents mutual interference. Additionally, we present a novel pipeline and filtering algorithm for extracting skeletal representations from various objects and release a high-quality dataset of 50,000 non-human pose-video pairs. Extensive quantitative and qualitative experiments demonstrate that PoseAnything outperforms the state-of-the-art methods and generalizes well across diverse subjects and poses.

{
    \small
    \bibliographystyle{ieeenat_fullname}
    \bibliography{arxiv}
}

\clearpage

\appendix



\clearpage

\section{Overview}
In this supplementary material, more details about the pro
posed PoseAnything method and more experimental results
 are provided, including:
 \begin{itemize}
     \item More Details about our dataset XPose (sec \ref{sec: Details about XPose})
     \item Details about attention weights based part matching in part-aware temporal coherence module (sec \ref{sec:details about part-aware temporal coherence module})
     \item details about motion decoupled CFG (sec \ref{sec:motion decoupled CFG})
     \item Ablation on Condition Injection Strategies (sec  \ref{sec: Ablation on Condition Injection Strategies})
     \item Ablation on Sparse Pose Condition Injection (sec \ref{sec: Ablation on Sparse Pose Condition Injection})
     \item Analysis of effect of CFG on pose-guided video generation of PoseAnything(sec \ref{sec:effect of CFG})
      \item more generation results of our PoseAnything (sec \ref{sec:more result})
\end{itemize}

More videos and comparison results are available on the provided \textbf{project page} and \textbf{demo video}. Please refer to them for further inspection.

\section{Details about XPose}
\label{sec: Details about XPose}
\begin{figure*}[h]
    \centering
    \includegraphics[width=0.9\linewidth]{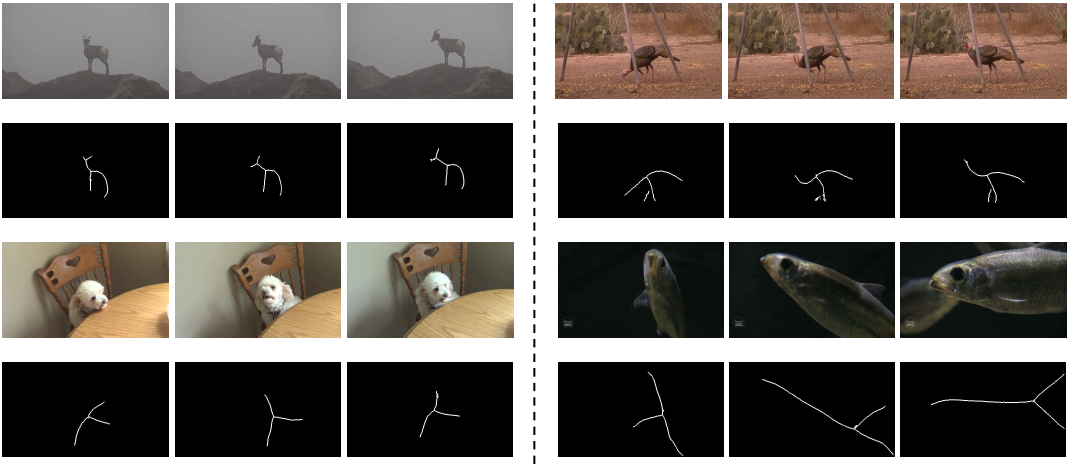}
    \vspace{-0.1in}
    \caption{Simple Subset of XPose}
    \vspace{-0.1in}
    \label{fig:Xpose easy example}
\end{figure*}

\begin{figure*}[h]
    \centering
    \includegraphics[width=0.9\linewidth]{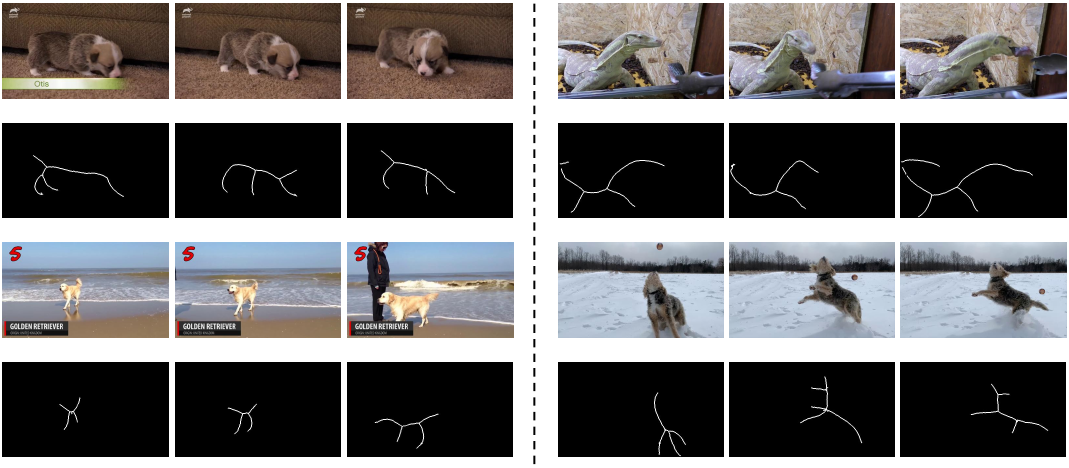}
    \vspace{-0.1in}
    \caption{Medium Subset of XPose}
    \vspace{-0.1in}
    \label{fig:Xpose medium example}
\end{figure*}

\begin{figure*}[h]
    \centering
    \includegraphics[width=0.9\linewidth]{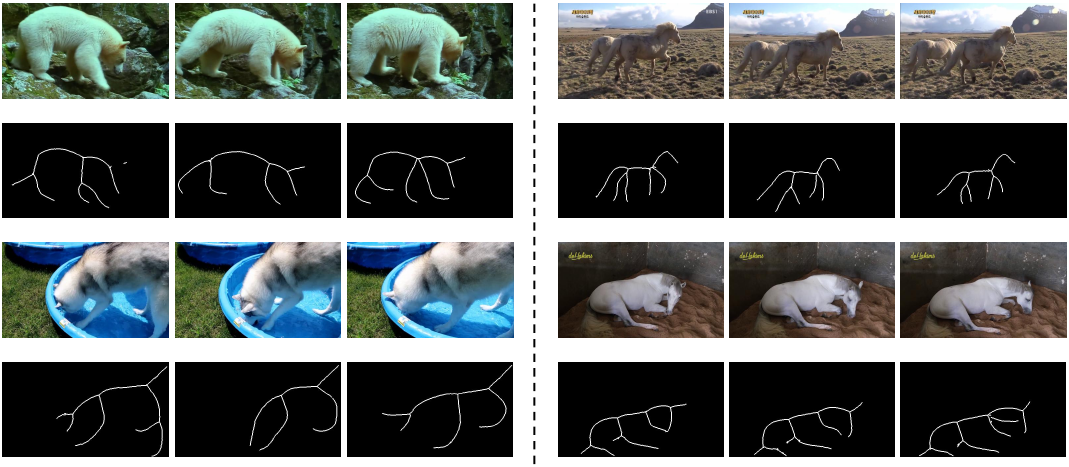}
    \vspace{-0.1in}
    \caption{Complex Subset of XPose}
    \vspace{-0.1in}
    \label{fig:Xpose complex example}
\end{figure*}

To effectively tackle the task of universal pose-guided video generation, which needs both robust generalization capabilities and precise pose-to-subject mapping, we curated a high-quality, non-human pose dataset termed \textbf{XPose}. As described in the main paper, the XPose dataset was generated from Koala \cite{wang2024koala36mlargescalevideodataset} and UltraVideo \cite{xue2025ultravideohighqualityuhdvideo} via a carefully designed pipeline coupled with a filtering algorithm to ensure its fidelity and diversity. In this section, we present a subset of XPose, offering a clear overview of its structural characteristics, complexity, and overall quality. To provide a more comprehensive and systematic demonstration, the dataset is stratified into three complexity levels according to the number of skeleton segments present: \textit{Simple} (1–3 segments), \textit{Medium} (4–6 segments), and \textit{Complex} (7–10 segments), which corresponds to Fig \ref{fig:Xpose easy example} \ref{fig:Xpose medium example} \ref{fig:Xpose complex example}, respectively.

This visualization highlights two critical features of XPose: (1) \textbf{Diversity}: $\text{XPose}$ exhibits a high degree of diversity, encompassing a wide range of backgrounds and dynamic motions. The dataset includes scenes set in diverse environments (e.g., underwater, sky, wilderness, home interiors) and features varied movements (e.g., twisting heads, running, swimming). (2) \textbf{High Fidelity} of Skeletons: The visualization confirms the high fidelity of the extracted skeletons, ensuring accurate pose representation. The high diversity in subject poses and high quality of the extracted skeletons is crucial ensures that the model learns accurate, noise-resistant pose-to-subject mapping, achieving the necessary robustness and high-fidelity generation required for the universal pose-guided video generation task. It provides strong support not only for our current universal generation method but also for future research in related work.

\section{Details about Part-aware Temporal Coherence Module}
\label{sec:details about part-aware temporal coherence module}
As detailed in the main paper, we designed a part-aware temporal coherence module for ensuring fine-grained inter-frame consistency. It is accomplished by decomposing the subject into distinct parts, establishing correspondences between identical parts across different frames, and enforcing these correspondences through cross-attention between matched part pairs. In the part-aware coherence module, the mechanism of part-matching is realized through the attention weights within the attention layers of the DiT \cite{peebles2023scalablediffusionmodelstransformers} blocks according to the formula:
\begin{equation}
s_{ij'} \thicksim s_{0j}\iff j'=\mathop{\arg\max}\limits_{t} \ \text{attn\_weight}[m_{0j}][m_{it}].
\end{equation}
Specifically, the underlying principle dictates that the attention weights between corresponding parts across different frames should be significantly higher than those between non-corresponding parts. However, the diffusion process inherently involves multiple diffusion timesteps ($t$), and the model's primary denoise focus and priorities shift significantly as the signal-to-noise ratio changes across these timesteps. For instance, at higher noise levels, the model tends to prioritize global structural features, whereas at lower noise levels, the focus shifts towards denoising and refining fine-grained details. Additionally, each forward pass traverses approximately 30 DiT blocks, with each block attending to different domains of information.
\begin{figure*}[h]
    \centering
    \includegraphics[width=0.9\linewidth]{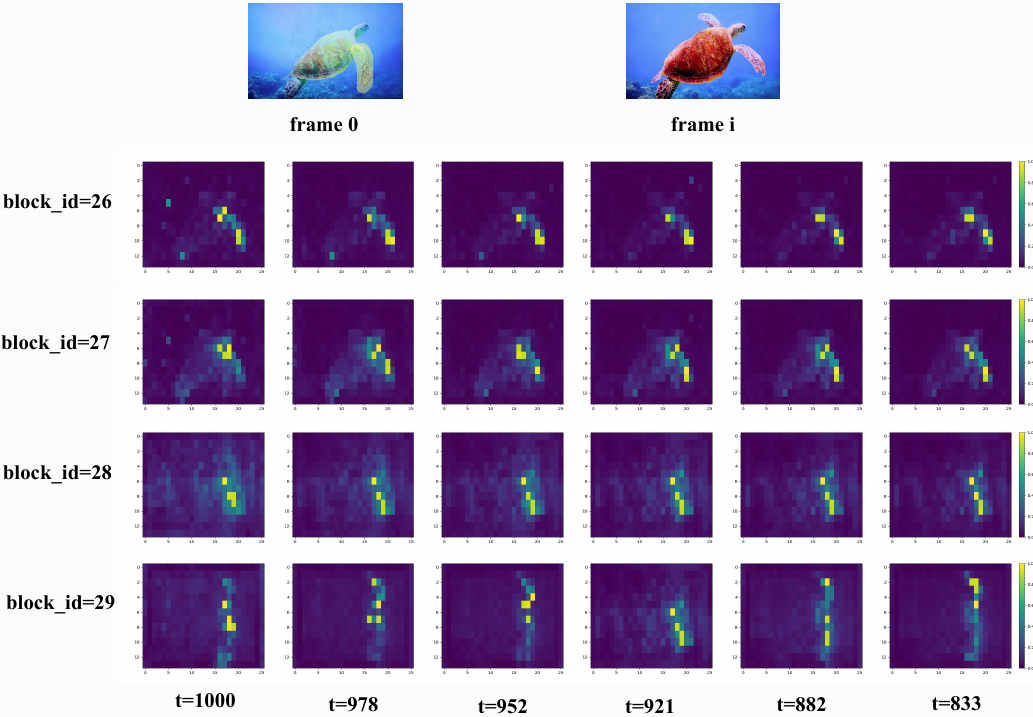}
    \caption{Visualization of attention weights}
    \vspace{-0.1in}
    \label{fig:attention weights}
\end{figure*}

To accurately achieve the desired part-level matching through attention weights, we conducted a systematic visualization of attention weights across various timesteps and different depths (DiT block indices) within the network. An illustrative example of these visualizations is presented in Fig.~\ref{fig:attention weights}. In this example, to obtain the visualization results of the attention weights, we designated the region corresponding to the sea turtle’s flipper (highlighted in yellow) in the first frame to compute the query vector ($q$), while the tokens from the corresponding regions in subsequent frames were used to compute the key and value vectors ($k, v$).For clarity of presentation, we only selected the token corresponding to $frame_{i}$ to compute $k$ and $v$. The resulting attention weights were averaged across the head dimension and then reshaped into an $h \times w$ spatial map.

The visualizations consistently demonstrate that the flipper region in subsequent frames appears significantly brighter, indicating that the token corresponding to the flipper in the first frame attends more strongly to the flipper region in subsequent frames. This observation is in accordance with our part-matching principle. Furthermore, this attention intensity difference is more pronounced at higher diffusion timesteps. Comparing attention weights across different block indices reveals that deeper blocks exhibit a more uniform attention distribution, whereas shallower blocks display more dispersed attention. Based on these findings, we select the attention weights from block whose id is 27 at timesteps greater than 975 as the basis for our part-matching mechanism.

\section{Details about Motion Decoupled CFG}
\label{sec:motion decoupled CFG}
In the main paper, we propose the subject-camera decoupled motion control CFG. By injecting the control conditions for subject and background motion respectively into the positive and negative anchors of CFG, we effectively prevent mutual interference between these two motion components. In this section, we provide a detailed explanation of the underlying principles behind this approach.

The proposed motion decoupling approach can be rigorously interpreted through the lens of optical flow and latent vector composition. In video generation, optical flow $\mathbf{F}_t$ characterizes the pixel-wise displacement between consecutive frames $I_t$ and $I_{t+1}$, encapsulating both subject and background dynamics:
\begin{equation}
    \mathbf{F}_t = \mathbf{F}_{\text{subject},t} + \mathbf{F}_{\text{camera},t}
\end{equation}
where $\mathbf{F}_{\text{subject},t}$ denotes subject-induced motion, and $\mathbf{F}_{\text{camera},t}$ represents camera-induced (background) motion.

By explicitly formulating the guidance vectors for subject and camera motion within the Classifier-Free Guidance (CFG) framework, we effectively decompose the overall optical flow field into independent, controllable components in the latent space. Let $\Delta\epsilon_{\text{subject}}$ and $\Delta\epsilon_{\text{camera}}$ be the guidance vectors corresponding to subject and camera motion, respectively. The overall latent noise guidance is then expressed as:
\begin{equation}
    \tilde{\epsilon} = \hat{\epsilon}_{\theta}(\mathbf{z}_t, \emptyset) + s_s \cdot \Delta\epsilon_{\text{subject}} - s_c \cdot \Delta\epsilon_{\text{camera}}
\end{equation}
where $s_s$ and $s_c$ are scalar weights controlling the strength of subject and camera guidance. Specifically, the subject motion guidance vector $\Delta\epsilon_{\text{subject}}$ enforces the desired subject trajectory as defined in the pose sequence, steering the latent noise prediction toward the target action. This can be formulated as:
\begin{equation}
    \Delta\epsilon_{\text{subject}} = \hat{\epsilon}_{\theta}(\mathbf{z}_t, \mathbf{c}_{\text{subject}}) - \hat{\epsilon}_{\theta}(\mathbf{z}_t, \emptyset)
\end{equation}
where $\mathbf{c}_{\text{subject}}$ denotes the subject pose condition.

Meanwhile, camera motion is mathematically equivalent to imposing a spatially uniform optical flow field $\mathbf{F}_{\text{camera},t}$ across the background, as a camera pan or tilt requires all background pixels to move coherently in the opposite direction of the camera's intended movement. In the latent space, this is encoded by:
\begin{equation}
    \Delta\epsilon_{\text{camera}} = \hat{\epsilon}_{\theta}(\mathbf{z}_t, \mathbf{c}_{\text{camera}}) - \hat{\epsilon}_{\theta}(\mathbf{z}_t, \emptyset)
\end{equation}
where $\mathbf{c}_{\text{camera}}$ specifies the camera motion condition.

The final guided motion prediction $\tilde{\epsilon}$ thus emerges as a superposition of these vectors (Eq.~2), analogous to the principle of vector addition in classical mechanics. Each motion component, subject-specific ($V_s$) and background/camera-induced ($V_{bg}$), contributes linearly to the aggregate optical flow. Notably, since the background control condition is injected into the negative anchor of the CFG, the resulting latent guidance vector for background motion $V_{bg}$, in the final merged result, is synthesized in a direction opposite to that indicated by the provided control condition. Mathematically, this relationship is captured by the final guided optical flow equation:
\begin{equation}
    \mathbf{F}_t = s_s V_s - s_c V_{bg}
\end{equation}
where $s_s$ and $s_c$ are the respective guidance strengths for subject and background, $V_s$ denotes subject motion, and $V_{bg}$ is the guidance vector derived from background control information. This principled design guarantees that the model correctly interprets and synthesizes both subject and background motion trajectories, thereby enabling disentangled and physically coherent video generation under joint control conditions.

This compositionality enables precise and independent control over both foreground and background movements, allowing for the joint or separate manipulation of subject actions and camera transitions via adjustment of the respective guidance strengths $s_s$ and $s_c$ within the CFG framework. Consequently, our approach facilitates flexible, disentangled video synthesis, where complex scene dynamics can be intuitively controlled by the user.

\section{Ablation on Condition Injection Strategies}
\begin{figure}[t]
    \centering
    \includegraphics[width=0.95\linewidth]{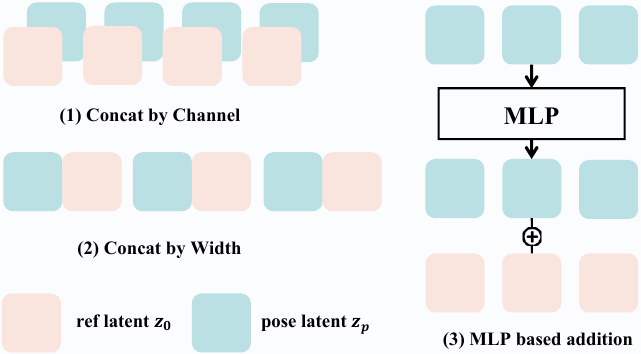}
    \caption{Different Condition Injection Strategies}
    \vspace{-0.1in}
    \label{fig:Injection Strategies}
\end{figure}

In the main paper, we propose three distinct strategies for incorporating pose condition into our framework: (1) \textbf{Concatenation by Channel} as \cite{bian2025gs} \cite{van2024generative}, (2) \textbf{Concatenation by Width} used in \cite{bai2024syncammastersynchronizingmulticameravideo}, and (3) \textbf{Multi-layer Perceptron (MLP)-based fusion}, in which the pose latent $Z_p$ is first processed by an MLP to align its shape with the reference latent $Z_0$, and subsequently added to $Z_0$. The architectural details of the three injection mechanisms are illustrated in Fig. \ref{fig:Injection Strategies}. To systematically compare the effectiveness of the proposed injection methods, we performed a dedicated ablation study, evaluating each approach both quantitatively and qualitatively on the TikTok dataset \cite{jafarian2021learning}. For this comparison, models incorporating only the respective injection strategies—excluding the Part-aware Temporal Coherence Module (PTCM)—were trained for 3,000 iterations. All models were trained using consistent hyper parameters: a batch size of $32$ and a learning rate of $2e-5$.

\label{sec: Ablation on Condition Injection Strategies}

\textbf{Quantitative results.} The following five metrics are employed to quantitatively compare the effectiveness of the three skeleton information injection strategies. For image-based metrics, the metric is first computed for each frame of a video, and the average across all frames is then taken as the final metric value for the video.
\begin{itemize}
    \item PSNR \cite{hore2010image}: The Peak Signal-to-Noise Ratio (PSNR) is one of the most prevalent and extensively utilized metrics for assessing image quality. A higher PSNR value indicates a superior quality of image reconstruction.
    \item SSIM \cite{wang2004image}: Structure Similarity Index Measure is derived from three aspects of image similarity: luminance, contrast and structure, based on the idea that the pixels have strong inter-dependencies especially when they are spatially close. The higher the SSIM score is, the more similar the two images are.
    \item L1: The L1 metric, quantifies the average absolute difference between the predicted and reference images. A lower L1 value signifies a closer resemblance between the reconstructed image and the ground truth.
    \item LPIPS \cite{zhang2018unreasonable}: The Learned Perceptual Image Patch Similarity (LPIPS) metric measures perceptual similarity between images using deep neural network features. Lower LPIPS scores indicate higher perceptual similarity, reflecting how closely the generated image aligns with human visual perception.
    \item FVD \cite{unterthiner2018towards} : The Fréchet Video Distance (FVD) is a widely adopted metric for evaluating the quality of generated videos. It compares the distributions of real and generated videos in a feature space, with lower FVD values indicating greater superior video generation quality.
\end{itemize}

The experimental results clearly demonstrate that the \textbf{Concatenation by Channel} approach consistently achieves superior performance across all evaluated metrics. Specifically, this method yields the highest perceptual quality, as evidenced by a $\text{PSNR}$ of $\mathbf{31.50}$ and an $\text{SSIM}$ of $\mathbf{0.8362}$. In addition, it attains a substantially lower reconstruction error, with an $\mathbf{L1}$ value of $\mathbf{2.79 \times 10^{-5}}$, corresponding to a $29.7\%$ reduction compared to the MLP method ($3.97 \times 10^{-5}$). The perceptual similarity, measured by $\text{LPIPS}$, is also highest for the Concatenation by Channel method, achieving the lowest score of $\mathbf{0.224}$. Most notably, with respect to temporal coherence, this method significantly outperforms all other strategies, achieving an $\text{FVD}$ of $\mathbf{133.95}$, a $52.8\%$ reduction relative to the next best method, Concat by MLP, which yields an $\text{FVD}$ of $283.79$. These results collectively highlight the effectiveness of the Concatenation by Channel strategy for skeleton information injection.

\begin{table}[h]
    \centering
     \vspace{-0.1in}
    \caption{Quantitative Comparison of Injection Strategies. }
    \vspace{-0.1in}
     \resizebox{\columnwidth}{!}{
    \begin{tabular}{lccccc} 
        \toprule
        Method & PSNR$\uparrow$ & SSIM$\uparrow$ & L1$\downarrow$ & LPIPS$\downarrow$ & FVD$\downarrow$ \\
        \midrule
        Concat by width & 29.16 & 0.7042 & 7.25E-05 & 0.370 & 415.13 \\
        Concat by MLP & 30.92 & 0.7829 & 3.97E-05 & 0.278 & 283.79 \\
         \rowcolor{gray!10}
        \textbf{Concat by channel}       & \textbf{31.50} & \textbf{0.8362} & \textbf{2.79E-05} & \textbf{0.224} & \textbf{133.95} \\
        \bottomrule
  \end{tabular}}
  \vspace{-0.1in}
    \label{tab:ablation on injection}
\end{table}

\begin{figure}[h]
    \centering
    \includegraphics[width=0.97\linewidth]{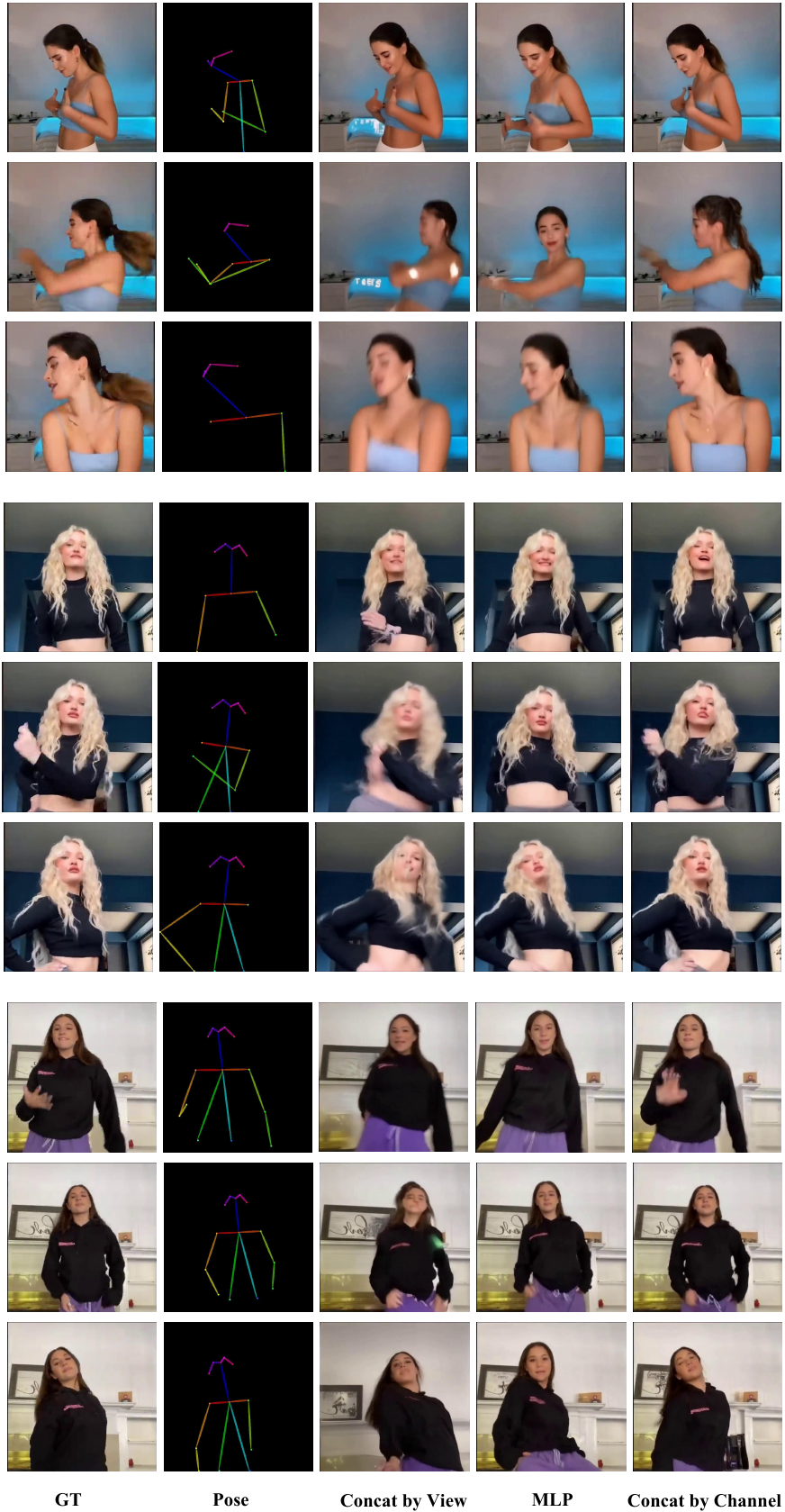}
    \caption{Qualitative Comparison of Different Injection Strategies}
    \vspace{-0.1in}
    \vspace{-0.1in}
    \label{fig:abalation on injection}
\end{figure}

\textbf{Qualitative Results.} As shown in Fig \ref{fig:abalation on injection}, the qualitative results indicate that the concat-by-channel approach produces more stable outputs compared to concat-by-width and MLP-based addition methods. Specifically, this is reflected in improved consistency of human motion, appearance uniformity, and overall image quality. Based on the aforementioned qualitative and quantitative comparison results, we adopt the concat-by-channel approach as our skeletal injection method at the coarse granularity level.

\section{Sparse Pose Condition Injection}
\label{sec: Ablation on Sparse Pose Condition Injection}
Current state-of-the-art human pose-guided video generation methods, including UniAnimate \cite{wang2024unianimatetamingunifiedvideo}, Animate-X \cite{tan2024animatexuniversalcharacterimage}, and trajectory-control approaches such as ATI~\cite{wang2025atitrajectoryinstructioncontrollable}, Tora~\cite{zhang2025toratrajectoryorienteddiffusiontransformer}, and SG-I2V~\cite{namekata2025sgi2vselfguidedtrajectorycontrol}, typically mandate dense control injection, requiring cues at every frame or across a substantial portion (i.e., $\geq 50\%$) of the sequence. This dense dependency contrasts with real-world applications where crucial control information is often concentrated within only a few \textbf{key frames}. This inherent disparity motivates us to enhance our model's capacity to learn effective temporal dynamics and motion propagation from \textbf{sparse pose injection}.

To cultivate this robust generalization capability, we devised a systematic \textbf{sparse sampling strategy} during training. Given the standard sequence length of 81 pose images, the dataloader randomly masks the input pose sequence according to a defined probability distribution for the remaining frames: (1) Dense Subset: $21$ to $81$ frames remain, applied with a probability of $35\%$, (2) Medium Subset: $11$ to $21$ frames remain, applied with a probability of $20\%$, (3) Sparse Subset: $1$ to $11$ frames remain, applied with a probability of $45\%$. Following the determination of the number of remaining control frames, we introduce further spatial variance by applying a \textbf{random masking scheme} with a $50\%$ probability, and a \textbf{uniform masking scheme} with a $50\%$ probability. Through the strategic integration of this stratified sampling approach, we aim to significantly bolster the model's temporal predictive capacity and improve its resilience and generalization under conditions of highly sparse skeleton input.

To evaluate the model's generalization ability under sparse pose conditions, we conduct experiments on the XPose-Benchmark described in the main paper. Specifically, we inject skeleton images at different sparsity levels,$\mathbf{100\%}$ (full condition), $\mathbf{50\%}$, $\mathbf{20\%}$, $\mathbf{10\%}$, $\mathbf{5\%}$, and $\mathbf{2.5\%}$, and compare the quantitative results.

\begin{figure}[h]
    \centering
    \includegraphics[width=0.9\linewidth]{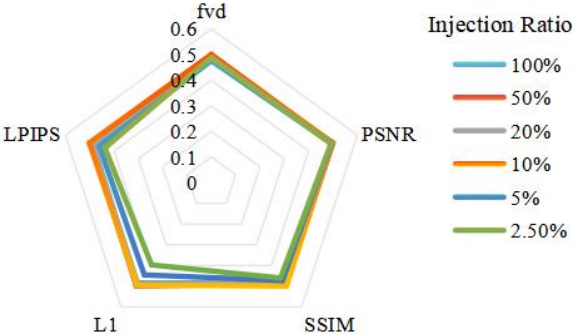}
    \caption{Visualization of Sparse Injection Comparison}
    \vspace{-0.1in}
    \label{fig:ablation on sparse injection}
\end{figure}

\textbf{Quantitative Results.} Tab.~\ref{tab:ablation on sparse injection} and Fig.~\ref{fig:ablation on sparse injection} summarize the quantitative comparison of model performance under different pose condition injection ratios. The results show that our model consistently maintains high performance across a broad range of injection densities, with only slight degradation observed even under conditions of extreme sparsity. Specifically, when the injection ratio is reduced from full conditioning ($100\%$) to $10\%$, all major image quality metrics ($\text{PSNR}$, $\text{SSIM}$, $\text{L1}$, $\text{LPIPS}$) experience negligible declines, and the temporal metric $\text{FVD}$ is even marginally improved, suggesting strong temporal coherence. Furthermore, under the most challenging scenario ($2.5\%$ injection), our model still delivers competitive results: while $\text{SSIM}$ and $\text{LPIPS}$ show the largest drops, $\text{PSNR}$ remains high and $\text{FVD}$ stays close to the baseline value. These observations highlight the robustness of our architecture in effectively interpolating and generalizing pose information across unconditioned frames. Such resilience under sparse pose injection conditions indicates the practical value of our method, especially for real-world applications where pose inputs may be infrequent or irregular.

\begin{table}[t]
    \centering
    \caption{Quantitative Result of Sparse Pose Condition Injection}
     \resizebox{\columnwidth}{!}{
    \begin{tabular}{lccccc} 
        \toprule
        Ratio & PSNR$\uparrow$ & SSIM$\uparrow$ & L1$\downarrow$ & LPIPS$\downarrow$ & FVD$\downarrow$ \\
        \midrule
        $100\%$ & 30.29 & 0.7114 & 8.19E-06 & 0.324 & 99.97 \\
        $50\%$ & 30.31 & 0.7145 & 8.03E-06 & 0.323 & 101.1 \\
        $20\%$ & 30.32 & 0.7097 & 8.08E-06 & 0.327 & 100.88 \\
        $10\%$ & 30.21 & 0.73334 & 8.13E-06 & 0.317 & 97.02 \\
        $5\%$ & 29.95 & 0.6934 & 9.01E-06 & 0.344 & 102.06 \\
        $2.5\%$ & 29.82 & 0.6757 & 1.01E-05 & 0.363 & 99.23 \\
        \bottomrule
  \end{tabular}}
  \vspace{-0.1in}
  \vspace{-0.1in}
    \label{tab:ablation on sparse injection}
\end{table}

\section{Effect of CFG in Pose-Guided Generation}
\label{sec:effect of CFG}
Classifier-Free Guidance (CFG) is a crucial technique in conditional diffusion models, designed to enhance the alignment between generated outputs and specified conditioning information (e.g., text, image, or pose). CFG operates by linearly extrapolating the predicted noise ($\hat{\epsilon}$) away from the unconditional estimate ($\hat{\epsilon}_{\emptyset}$) toward the conditional estimate ($\hat{\epsilon}_{\theta}$), controlled by a guidance scale parameter $s$:
\begin{equation}
    \tilde{\epsilon} = \hat{\epsilon}_{\theta}(\mathbf{z}_t, \mathbf{c}) + s \cdot \left( \hat{\epsilon}_{\theta}(\mathbf{z}_t, \mathbf{c}) - \hat{\epsilon}_{\theta}(\mathbf{z}_t, \emptyset) \right)
\end{equation}
Within the context of diffusion-based video generation, the CFG scale $s$ serves as a hyperparameter that governs the degree of fidelity to the motion condition. Increasing $s$ generally enforces stricter adherence to the input pose sequence, resulting in sharper and more pronounced movements, albeit sometimes at the expense of sample diversity and generation stability.

To empirically evaluate the performance of PoseAnything to the strength of conditional guidance, we introduced CFG into PoseAnything and conducted a series of experiments. Specifically, as a negative anchor, we injected a pose sequence in which every valid pose was identical to the first frame's pose. The primary objective was to systematically characterize model performance across a range of CFG scale values, thereby elucidating the trade-off between pose fidelity and overall generation quality. Our experimental protocol encompassed two distinct sparsity settings: (1) Sparse: Pose conditions were injected into $10\%$ of the total generated frames; (2) Dense: Pose conditions were injected into $100\%$ of the generated frames.
\begin{figure*}[h]
    \centering
    \includegraphics[width=0.95\linewidth]{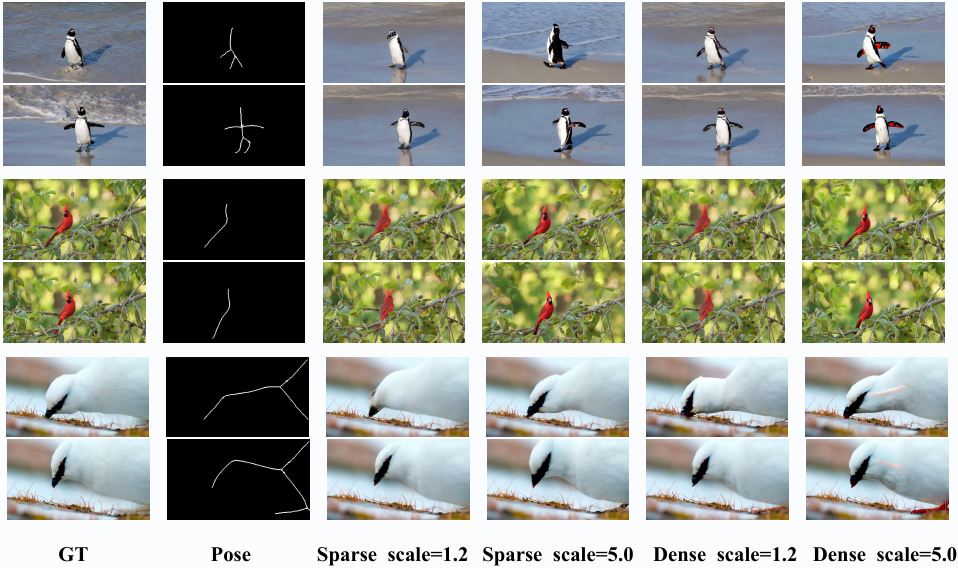}
    \caption{Qualitative Comparison of Different CFG Scale under Sparse/Dense Injection}
    \vspace{-0.1in}
    \label{fig:cfg}
\end{figure*}

\textbf{Quantitative results.} As observed from Tab ~\ref{tab:cfg}, increasing the CFG scale leads to a consistent degradation across all quantitative metrics, with this effect being more pronounced under the dense conditioning setting. Specifically, as the value of \textit{cfg scale} increases, both PSNR and SSIM decrease, while L1, LPIPS, and FVD scores increase, indicating a decline in both reconstruction fidelity and perceptual quality. This trend is especially evident when pose conditions are densely injected (\textit{dense} setting), suggesting that higher conditioning density amplifies the impact of excessive guidance strength. This phenomenon can be attributed to the fact that, as the density of input pose information increases, the conditional signal becomes more dominant. While moderate CFG scales help align generated outputs with the desired pose, excessively high CFG scales (e.g., 5.0) may over-constrain the model, reducing the diversity and stability of generated samples. This over-constraining effect manifests as lower PSNR and SSIM, along with elevated LPIPS and FVD scores, reflecting poorer visual quality and diminished sample diversity.

\begin{table}[h]
    \centering
    \vspace{-0.1in}
    \caption{Quantitative Result of Applying CFG}
    \vspace{-0.1in}
    \resizebox{\columnwidth}{!}{
    \begin{tabular}{lcclcccc} 
        \toprule
        Scale & Setting & PSNR$\uparrow$ & SSIM$\uparrow$ & L1$\downarrow$ & LPIPS$\downarrow$ & FVD$\downarrow$ \\
        \midrule
        5.0 & dense & 29.92 & 0.6768 & 9.77E-06 & 0.3471 & 100.22 \\
        3.0 & dense & 30.47 & 0.7126 & 7.82E-06 & 0.3241 & 99.78 \\
        1.5 & dense & 30.52 & 0.7172 & 7.67E-06 & 0.3226 & 99.94 \\
        1.2 & dense & 30.61 & 0.7245 & 7.42E-06 & 0.3192 & 99.29 \\
        \midrule 
        5.0 & sparse & 29.84 & 0.6730 & 9.64E-06 & 0.3557 & 100.83 \\
        3.0 & sparse & 30.14 & 0.6900 & 8.73E-06 & 0.3441 & 98.91 \\
        1.5 & sparse & 30.56 & 0.7122 & 7.49E-06 & 0.3276 & 99.11 \\
        1.2 & sparse & 30.68 & 0.7183 & 7.28E-06 & 0.3245 & 99.01 \\
        \bottomrule
    \end{tabular}}
    \vspace{-0.1in}
    \label{tab:cfg}
\end{table}

\textbf{Qualitative Results.} Qualitative results are presented in Fig. \ref{fig:cfg}. As observed, increasing the CFG scale enhances the model’s ability to generate the target pose, resulting in more accurate pose alignment. However, under strong skeletal conditioning—particularly in the dense injection setting—a higher CFG scale tends to introduce visual artifacts, such as unnatural limb shapes or distortions in body structure, which may negatively impact the overall visual quality.

In summary, the selection of CFG scale should be adapted to the density of pose conditioning. When the input pose density is high (i.e., dense setting), a lower CFG scale is preferable to prevent over-constraining the model and to maintain a balance between conditional fidelity and visual quality. Conversely, when the pose conditioning is sparse, increasing the CFG scale can effectively enhance the model’s ability to fit the provided pose information without significantly compromising generation stability or diversity. This adaptive strategy enables the model to achieve optimal performance across varying levels of conditioning strength, thereby providing practical guidance for tuning diffusion-based pose-guided generation systems.

\section{More generation results of PoseAnything}
\label{sec:more result}
\begin{figure*}[h]
    \centering
    \includegraphics[width=0.9\linewidth]{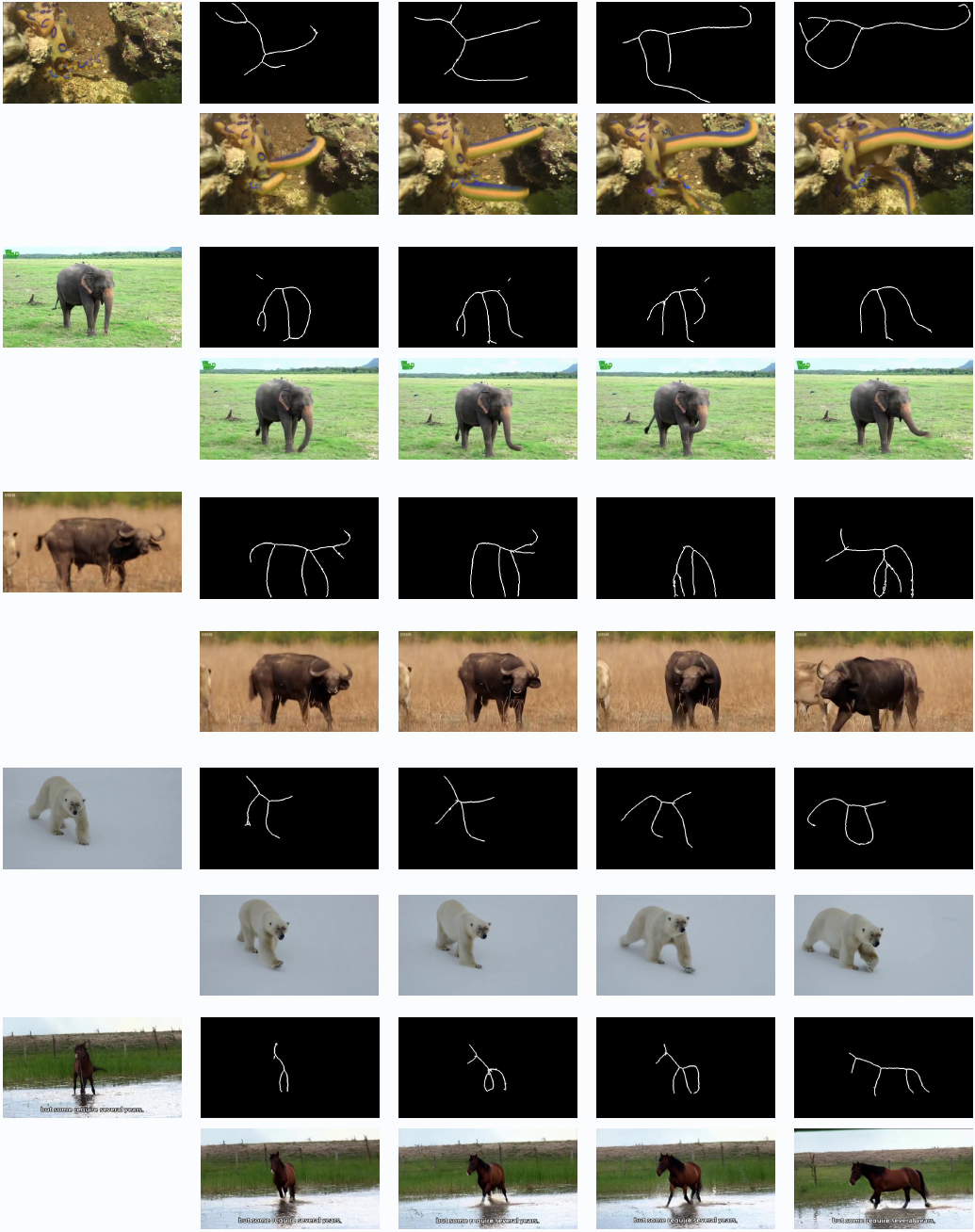}
    \caption{More Results of PoseAnything} 
    \label{fig:more results}
\end{figure*}

In this section, we present more generated samples to further demonstrate the effectiveness and robustness of our model. As illustrated in Fig.~\ref{fig:more results}, our approach PoseAnything consistently achieves precise pose adherence, accurately following the provided motion cues across diverse scenarios. Moreover, the results reveal a remarkable generalization capability, with the model maintaining high performance across a wide variety of scene backgrounds, subject appearances, and motion types. Importantly, our method preserves both visual and temporal coherence, ensuring smooth and realistic transitions throughout the generated sequences. These findings collectively underscore the reliability and adaptability of our model in handling complex and challenging motion generation tasks.

\begin{figure*}[h]
    \centering
    \includegraphics[width=0.9\linewidth]{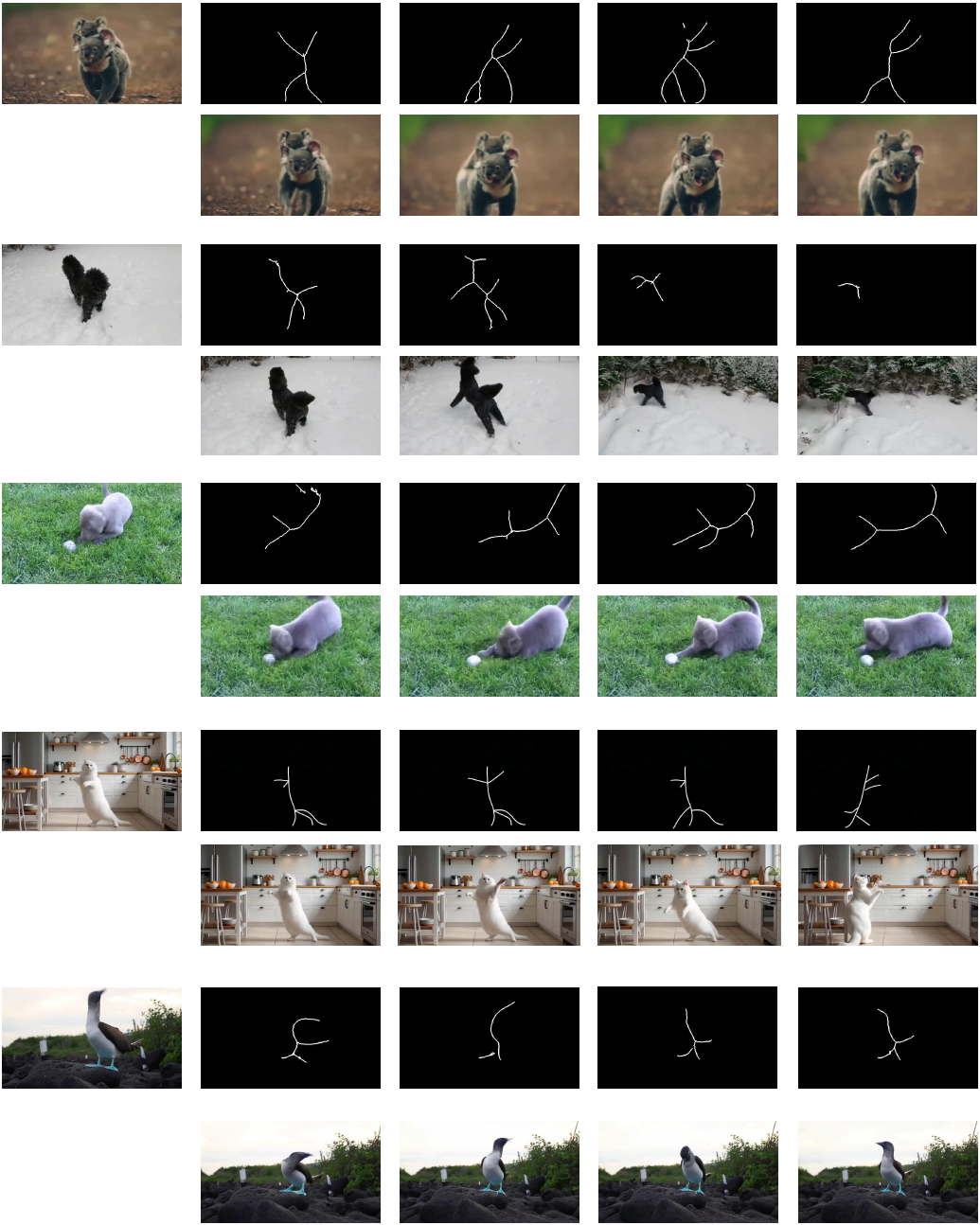}
\end{figure*}

\clearpage 

\end{document}